\pdfoutput=1
\documentclass[10pt,twocolumn,letterpaper]{article}

\usepackage[pagenumbers]{cvpr}

\usepackage{graphicx}
\usepackage{amsmath}
\usepackage{amssymb}
\usepackage{bm}          %
\usepackage{booktabs}
\usepackage{multirow}
\usepackage{array}
\usepackage{xcolor}
\usepackage{colortbl}    %
\usepackage{pifont}      %
\usepackage{placeins}    %
\usepackage{dblfloatfix}

\newcommand{\cmark}{\ding{51}}

\definecolor{best}{RGB}{198,222,227}
\definecolor{second}{RGB}{230,240,242}
\definecolor{cvprblue}{rgb}{0.21,0.49,0.74}
\usepackage[pagebackref,breaklinks,colorlinks,citecolor=cvprblue]{hyperref}
\graphicspath{{figures/}}
\AtBeginEnvironment{table}{\renewcommand{\arraystretch}{0.9}}
\AtBeginEnvironment{table*}{\renewcommand{\arraystretch}{0.9}}

\def\confName{CVPR}
\def\confYear{2027}

\newcommand{\method}{GradSplat\xspace}
\newcommand{\Kshape}{5\xspace}
\newcommand{\Kapp}{3\xspace}
\newcommand{\dB}{\,dB\xspace}

\newcommand{\rowname}[1]{\rotatebox{90}{\scriptsize #1}}

\providecommand{\mainNarrowOurs}{}\renewcommand{\mainNarrowOurs}{33.683}
\providecommand{\mainWideOurs}{}\renewcommand{\mainWideOurs}{26.755}
\providecommand{\mainNarrowAny}{}\renewcommand{\mainNarrowAny}{19.629}
\providecommand{\mainWideAny}{}\renewcommand{\mainWideAny}{15.525}
\providecommand{\groupsBase}{}\renewcommand{\groupsBase}{27.278}
\providecommand{\groupsShape}{}\renewcommand{\groupsShape}{34.207}
\providecommand{\groupsAll}{}\renewcommand{\groupsAll}{34.231}
\providecommand{\groupsNoCentres}{}\renewcommand{\groupsNoCentres}{34.505}
\providecommand{\poseOursNarrow}{}\renewcommand{\poseOursNarrow}{0.794}
\providecommand{\poseYonoNarrow}{}\renewcommand{\poseYonoNarrow}{0.593}
\providecommand{\zsDlOurs}{}\renewcommand{\zsDlOurs}{24.787}

\begin{document}

\title{\method: Gradients Inside the Forward Pass for\\
       Feed-Forward 3D Gaussian Splatting}

\author{%
Jingong Chen$^{1,2}$\quad Linlian Jiang$^{3,4,\dagger}$\quad Guangyuan Li$^{5,2,\dagger}$\quad Xinxin Zuo$^{3}$\\
Yang Wang$^{3,4}$\quad Chulwoo Pack$^{1}$\quad Kyle Gao$^{2}$\quad Kwanghee Won$^{1,*}$\\[6pt]
{\normalsize $^{1}$South Dakota State University\quad $^{2}$Aalto University\quad $^{3}$Concordia University}\\
{\normalsize $^{4}$Mila -- Quebec AI Institute\quad $^{5}$ELLIS Institute Finland\quad $^{\dagger}$Equal contribution\quad $^{*}$Corresponding author}\\[-6pt]
}

\makeatletter
\twocolumn[%
  \@maketitle
  \begin{center}
  {\includegraphics[width=\textwidth]{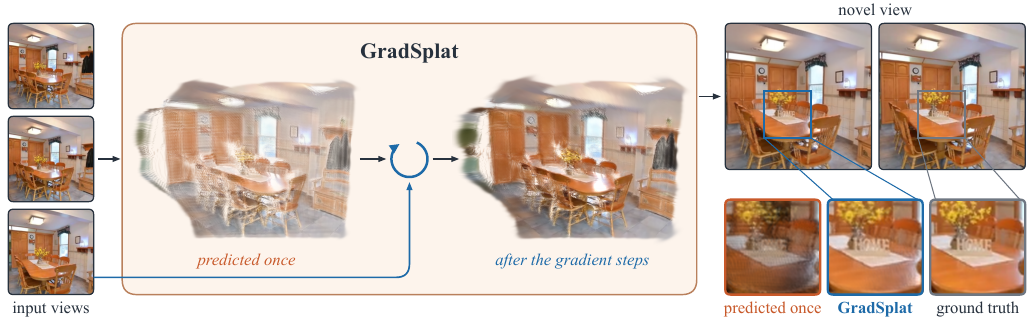}}
  \captionof{figure}{\textbf{\method} is a feed-forward model that reconstructs a scene as 3D Gaussians from a few views and refines them by gradient steps on the input views inside the model. Its Gaussians are rendered as predicted once and after the steps. One novel view is shown as rendered by \method and in the ground truth, and the details compare it before the steps, after them and in the ground truth.}
  \label{fig::teaser}
  \end{center}
  \vspace{2pt}
]
\makeatother

\begin{abstract}
Feed-forward 3D Gaussian Splatting predicts a scene from a few images, but our predictor is less accurate than its own Gaussians optimized per scene on those images.
This gap lies mostly in the shape of the Gaussians, their scale and rotation, since optimizing the shape alone recovers almost all of it and opacity and color recover less.
Gradient steps on the input views correct this shape, and the model holds these views at inference.
\textit{Log-anisotropy correction} changes the axis ratios of each Gaussian in log space at constant volume, and \textit{in-forward gradient refinement} renders the Gaussians into the input views and applies the gradient of the photometric loss with a learned recurrent update, first to opacity and color and then to the shape.
\textbf{\method} is a feed-forward model that refines its own Gaussians by learned gradient steps inside its forward pass, with no target view.
On both complete six-view RealEstate10K protocols, it has the best PSNR and SSIM among nine methods and keeps its accuracy from $6$ to $42$ input views.
Without fine-tuning, it is the most accurate on DL3DV and has the best PSNR and SSIM on ScanNet++.
The code will be released at \href{https://github.com/Jingong-Chen/GradSplat}{\textcolor{cvprblue}{\mbox{https://github.com/Jingong-Chen/GradSplat}}}.
\end{abstract}

\setcounter{dbltopnumber}{1}
\section{Introduction}
\label{sec::intro}

Synthesizing novel views of a scene from a few images is a fundamental problem in computer vision.
Models built on 3D Gaussian Splatting~\cite{kerbl3Dgaussians}, such as pixelSplat~\cite{charatan2024pixelsplat}, MVSplat~\cite{chen2024mvsplat}, DepthSplat~\cite{xu2025depthsplat}, NoPoSplat~\cite{ye2025noposplat}, AnySplat~\cite{anysplat}, YoNoSplat~\cite{yonosplat} and others~\cite{zhang2025transplat,wewer2024latentsplat,zhang2024gslrm,wang2024freesplat,volsplat}, now solve it with a feed-forward network but predict the Gaussians once.

\begin{figure*}[t]
\centering
\includegraphics[width=\textwidth]{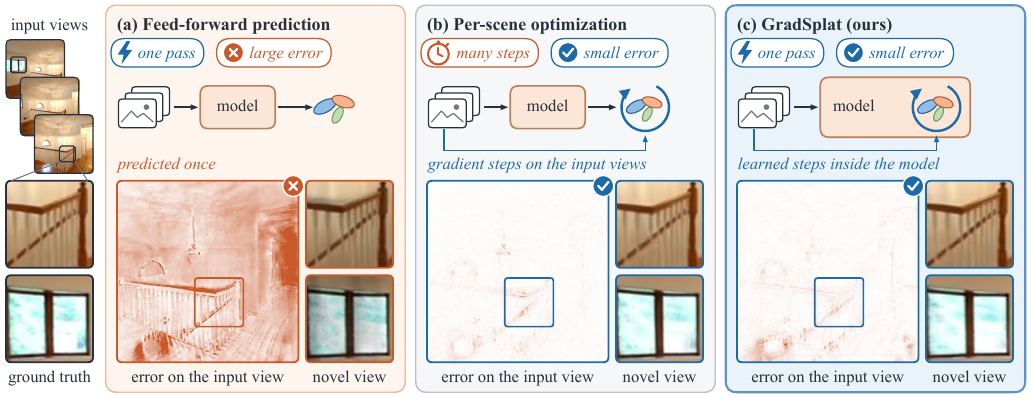}
\caption{\textbf{Motivation.} A feed-forward model predicts the Gaussians in one pass, but rendered back into an input view they leave a large error (a).
Per-scene optimization of their scale and rotation removes most of it, but takes $1500$ gradient steps on the input views (b).
\method takes gradient steps with a learned update inside the model and keeps the single pass (c).
The maps show the mean absolute error against the input view on one scale.
The small pictures show the marked window of the rendering and a window of a novel view.
The left column shows three input views and the same two windows in the input view and in the ground truth, and the box on the upper input view marks where the second window lies in the scene.
The model of (a) is our base predictor, and the scene is chosen for a visible difference.}
\label{fig::motiv}
\end{figure*}

Per-scene optimization~\cite{mildenhall2020nerf,barron2021mipnerf,barron2022mipnerf360,mueller2022instant,barron2023zipnerf,kerbl3Dgaussians,xie2026fact} fits the parameters to one scene by thousands of gradient steps.
Applied to the output of our base predictor on its input views, with no target view, it raises the novel-view PSNR from $\groupsBase$ to $\groupsAll$\dB on eight RealEstate10K~\cite{zhou2018stereo} scenes.
This is the amortization gap~\cite{cremer2018inference} of predicting the parameters with one network shared by all scenes.
Recent work extends feed-forward models by multi-view matching and depth priors~\cite{chen2024mvsplat,zhang2025transplat,xu2025depthsplat}, by a pose-free geometry backbone~\cite{ye2025noposplat,anysplat,yonosplat}, or by a recurrent network that updates the Gaussians from the rendering error~\cite{resplat}.
The first two keep a single prediction of the Gaussians.
The third computes no gradient and updates all Gaussian parameters.

We find that the gap is carried by one parameter group, the shape of the Gaussians, which is the scale and rotation that form their covariance.
Optimized alone on the input views, the shape recovers almost all of the gap, whereas opacity and color recover less and freezing the centers loses nothing.
The predicted Gaussians are less anisotropic than the optimized ones, and they do not reproduce even their own input views (\cref{fig::motiv}(a)).
Optimizing their scale and rotation removes most of this error and improves the novel views, but takes $1500$ gradient steps (\cref{fig::motiv}(b)).
A feed-forward model could therefore close the gap itself, if it learned to take gradient steps inside its forward pass.

We present \method, which takes learned gradient steps inside the feed-forward model (\cref{fig::motiv}(c)), first on opacity and color and then on the shape.
\textit{Log-anisotropy correction} changes the axis ratios of each Gaussian in log space at constant volume and adds a rotation increment.
\textit{In-forward gradient refinement} renders the Gaussians into the input views, takes the gradient of the photometric loss through the rasterizer, and applies it with a learned recurrent update.
The steps are fixed in number, unrolled in training and read only the input views.
\method therefore remains a feed-forward model, as in \cref{fig::teaser}, and uses no target view at inference.
On both complete six-view RealEstate10K protocols, \method has the best PSNR and SSIM among nine evaluated methods, and one model keeps its accuracy from $6$ to $42$ input views.
Without fine-tuning, it is the most accurate of these methods on DL3DV~\cite{ling2024dl3dv}.
Without input poses, it runs on its predicted cameras at lower accuracy.

Our contributions are threefold:
\begin{itemize}[leftmargin=*]
\itemsep2pt
\item We identify that the amortization gap of feed-forward 3D Gaussian Splatting is carried by the shape of the Gaussians, and that gradient steps on the input views, available at inference, correct it.

\item \method, a feed-forward model that refines its own Gaussians by learned gradient steps on the input views inside its forward pass, with no target view, through two modules, \textit{log-anisotropy correction} and \textit{in-forward gradient refinement}.

\item State-of-the-art PSNR and SSIM on both complete six-view RealEstate10K protocols among nine evaluated methods, with one model serving $6$ to $42$ input views.
\end{itemize}

\section{Related Work}
\label{sec::related}
\begin{figure*}[t]
\centering
\includegraphics[width=\textwidth]{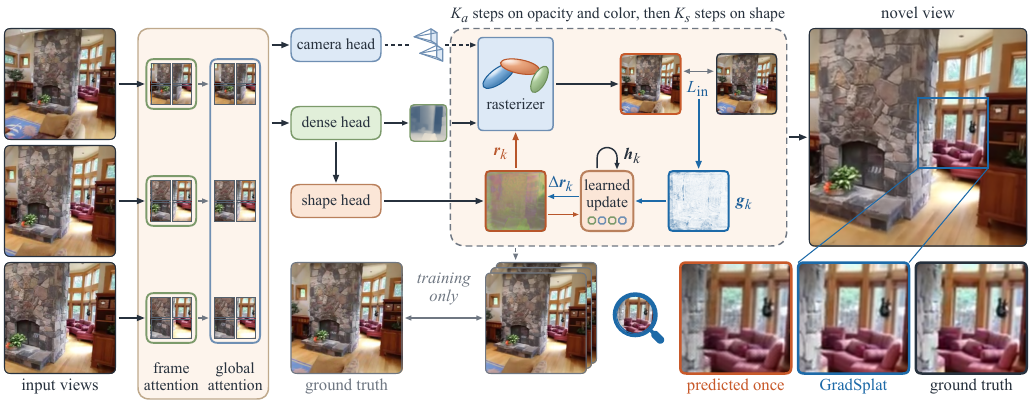}
\caption{\textbf{Overview.} A backbone feeds a camera head, a dense head and a shape head.
Each refinement step renders the Gaussians into the input views and applies the learned update $\mathcal{U}_\psi$ to the gradient $\bm{g}_k$ of $\mathcal{L}_{\mathrm{in}}$.
The four marks in $\mathcal{U}_\psi$ are its transformer blocks, with attention within a view (green) and over all views (blue).}
\label{fig::method}
\end{figure*}

\subsection{Feed-Forward 3D Gaussian Splatting}

Radiance fields~\cite{mildenhall2020nerf,mueller2022instant,fridovich2022plenoxels,chen2022tensorf,barron2022mipnerf360,barron2023zipnerf} and 3D Gaussian Splatting~\cite{kerbl3Dgaussians} are optimized per scene, whereas generalizable renderers~\cite{yu2021pixelnerf,wang2021ibrnet,chen2021mvsnerf,liu2022neuray,suhail2022gpnr,varma2023gnt} learn a network across scenes.
Feed-forward 3DGS adopts this paradigm. Posed methods mainly differ in how they place the Gaussian centers: pixelSplat~\cite{charatan2024pixelsplat} samples them from a predicted distribution, MVSplat~\cite{chen2024mvsplat} uses a plane-sweep cost volume, and TranSplat~\cite{zhang2025transplat} and DepthSplat~\cite{xu2025depthsplat} add monocular depth priors.
Others change the decoder or view aggregation: GS-LRM~\cite{zhang2024gslrm} decodes per-pixel Gaussians from transformer tokens, latentSplat~\cite{wewer2024latentsplat} renders feature Gaussians through a generative decoder, FreeSplat~\cite{wang2024freesplat} fuses Gaussians across overlapping views, and VolSplat~\cite{volsplat} predicts one Gaussian per occupied voxel.

Without input poses, Splatt3R~\cite{smart2024splatt3r} extends the pointmaps~\cite{wang2024dust3r} of MASt3R~\cite{leroy2024mast3r} with Gaussian parameters, NoPoSplat~\cite{ye2025noposplat} predicts Gaussians in the camera frame of one input view, AnySplat~\cite{anysplat} builds on VGGT~\cite{wang2025vggt} to also predict camera parameters, and YoNoSplat~\cite{yonosplat} predicts per-view Gaussians and poses, handling posed and pose-free input with one model.
However, all these methods predict the Gaussians in a single forward pass, leaving prediction errors uncorrected at inference.

\subsection{Refining a Predicted Reconstruction}
One line corrects reconstructions through images. NeRFLiX~\cite{zhou2023nerflix} restores rendered views with an inter-viewpoint mixer, GANeRF~\cite{roessle2023ganerf} optimizes the radiance field with a discriminator on rendered patches, and Difix3D+~\cite{difix3d} removes artifacts from rendered views with a single-step diffusion model before distilling them into the reconstruction.
Generative priors~\cite{jiang2026trace} regularize or replace reconstruction: ReconFusion~\cite{wu2024reconfusion} regularizes a NeRF with a diffusion prior, CAT3D~\cite{gao2024cat3d} reconstructs from views generated by a multi-view diffusion model, and ZeroNVS~\cite{sargent2024zeronvs} applies a 3D-aware diffusion model to a single image.
These methods operate on rendered or generated images rather than updating the representation directly from the input views.

Another line learns the optimization itself. LISTA~\cite{gregor2010learning} unfolds a fixed number of solver iterations into a trained network, a learned optimizer~\cite{andrychowicz2016learning} maps gradients to parameter updates, RAFT~\cite{teed2020raft} updates a flow field with a recurrent unit, and deep equilibrium models~\cite{bai2019deq} solve for the fixed point of a weight-tied network.
In view synthesis, DeepView~\cite{flynn2019deepview} builds multiplane images by learned gradient descent, G3R~\cite{chen2024g3r} updates large-scene representations from differentiable-rendering gradients, and QuickSplat~\cite{liu2025quicksplat} learns Gaussian updates and densification from rendering gradients for surface reconstruction.
However, these target multiplane images, large driving and drone scenes, or indoor surfaces rather than feed-forward 3DGS.
Closest to ours, ReSplat~\cite{resplat} updates the Gaussians of a feed-forward model from input-view rendering errors with a gradient-free recurrent network.
We instead take learned gradient steps on input views inside the feed-forward model, updating opacity and color before shape.

\section{Method}
\label{sec::method}

As shown in~\cref{fig::method}, \method predicts scene Gaussians with a feed-forward network and refines them in the same forward pass.
The refinement takes learned gradient steps on input views, updating appearance first and then shape through a log-anisotropy correction.

\subsection{Gaussians from a Feed-Forward Model}
\label{sec::arch}

The input is $V$ views, each an image $\bm{I}_v\in\mathbb{R}^{H\times W\times3}$ of height $H$ and width $W$ with a camera $\pi_v$, its pose and intrinsics.
The output is $VHW$ Gaussians, one per input pixel.
Each Gaussian has a center $\bm{\mu}\in\mathbb{R}^3$, a scale $\bm{s}\in\mathbb{R}_{+}^{3}$ of three axis lengths, a rotation $\bm{R}\in SO(3)$, an opacity $\alpha\in(0,1)$ and spherical-harmonic color coefficients $\bm{c}\in\mathbb{R}^{12}$.
Its covariance is $\bm{\Sigma}=\bm{R}\,\mathrm{diag}(\bm{s})^2\bm{R}^{\top}$.
We call $(\bm{s},\bm{R})$ the shape and $(\alpha,\bm{c})$ the appearance of a Gaussian, and write $\theta$ for the parameters of all Gaussians.
A differentiable rasterizer $\mathcal{R}(\theta,\pi)$~\cite{kerbl3Dgaussians} renders them from a camera $\pi$, with colors clamped to $[0,1]$.

Our backbone is a geometry transformer with alternating frame and global attention~\cite{wang2025vggt}, initialized from VGGT-$\Omega$~\cite{wang2026vggtomega}.
A camera head predicts the camera of every view.
A dense head predicts a depth for every input pixel.
It decodes the other parameters of the Gaussian of that pixel from a feature map $\bm{F}$ at half the image resolution.
With input poses, the depth is multiplied by the ratio of the given to the predicted distance between the first two cameras.
Without input poses, the predicted cameras serve as $\pi_v$, and a constant $\kappa$ shared by all scenes replaces this ratio and multiplies their translations.
The center $\bm{\mu}$ is the pixel unprojected through $\pi_v$ at this depth, plus a bounded learned offset.

Per-scene optimization starts from the predicted Gaussians and minimizes mean squared error on the input views,
\begin{equation}
\label{eq::perscene}
\mathcal{L}_{\mathrm{in}}(\theta)=\frac{1}{3VHW}\sum_{v=1}^{V}\bigl\lVert\mathcal{R}(\theta,\pi_v)-\bm{I}_v\bigr\rVert_2^2 .
\end{equation}
\method takes its gradient steps on this loss.

\subsection{Log-Anisotropy Correction}
\label{sec::shape-head}

The log-anisotropy correction changes the axis ratios of a predicted Gaussian at constant volume and adds a rotation increment.
The dense head outputs a raw scale $\bm{\rho}\in\mathbb{R}^3$ and a quaternion for every Gaussian.
The normalized quaternion gives the base rotation $\bar{\bm{R}}$, and the base scale is
\begin{equation}
\label{eq::base}
\bar{\bm{s}}=\tau\,\mathrm{softplus}(\bm{\rho}),
\end{equation}
with $\tau=10^{-3}$ a fixed unit of length.
A convolutional shape head reads $\bm{F}$ and the Sobel responses of the input image and predicts a raw output $\bm{r}^{\delta}\in\mathbb{R}^3$ for the scale and $\bm{r}^{\omega}\in\mathbb{R}^3$ for the rotation of every Gaussian.
It corrects the scale in log space and the rotation by an axis-angle increment in the frame of the Gaussian,
\begin{equation}
\label{eq::aniso}
\begin{aligned}
\bm{\delta}&=4\,\bm{P}\tanh(\bm{r}^{\delta}), &
\bm{s}&=\min\bigl(\bar{\bm{s}}\odot e^{\bm{\delta}},\,0.3\bigr),\\
\bm{\omega}&=\tanh(\bm{r}^{\omega}), &
\bm{R}&=\bar{\bm{R}}\exp\bigl([\bm{\omega}]_\times\bigr),
\end{aligned}
\end{equation}
with $\bm{\delta}\in\mathbb{R}^3$ the log-anisotropy, $\bm{\omega}\in\mathbb{R}^3$ the rotation increment, $[\cdot]_\times$ the skew-symmetric matrix of a vector, $\bm{1}\in\mathbb{R}^3$ the vector of ones and $\bm{P}=\mathrm{diag}(\bm{1})-\tfrac13\bm{1}\bm{1}^{\top}$ the projector onto zero-sum vectors.
The minimum with $0.3$ bounds every axis length.
The product $\odot$, $e^{(\cdot)}$, $\tanh$ and $\min$ act per element, and $\exp$ is the matrix exponential.

\noindent\textbf{Volume preservation.} Since $\bm{1}^{\top}\bm{P}=\bm{0}^{\top}$, the correction keeps $\bm{1}^{\top}\bm{\delta}=0$.
With the base covariance $\bar{\bm{\Sigma}}$, this gives
\begin{equation*}
\det\bm{\Sigma}=e^{2\,\bm{1}^{\top}\bm{\delta}}\det\bar{\bm{\Sigma}}=\det\bar{\bm{\Sigma}},\qquad\bar{\bm{\Sigma}}=\bar{\bm{R}}\,\mathrm{diag}(\bar{\bm{s}})^2\bar{\bm{R}}^{\top},
\end{equation*}
wherever the bound $0.3$ is inactive.
The log ratio of two axes changes by the difference of their components of $\bm{\delta}$, so the correction moves extent between axes at constant volume.

\noindent\textbf{Range.} Every component of $4\tanh(\bm{r}^{\delta})$ lies in $(-4,4)$, and $\bm{P}$ subtracts the same mean from each.
Two components of $\bm{\delta}$ therefore differ by less than $8$, so the correction changes an axis ratio by a factor below $e^{8}$.
The angle $\lVert\bm{\omega}\rVert_2$ of the rotation increment is below $\sqrt{3}$.

\noindent\textbf{Initialization.} The last layer of the shape head is initialized to zero, so $\bm{\delta}=\bm{0}$ and $\bm{\omega}=\bm{0}$.
Adding the head therefore leaves the output of the model unchanged.

\begin{figure*}[t]\centering\small
\setlength{\tabcolsep}{0.5pt}\renewcommand{\arraystretch}{0.8}
\begin{tabular}{@{}c@{\hspace{2.5pt}}w{c}{0.1360\textwidth}*{9}{w{c}{0.0907\textwidth}}@{}}
 & Inputs & pixelSplat & MVSplat & TranSplat & DepthSplat & VolSplat & ReSplat & YoNoSplat & Ours & GT \\
\rotatebox{90}{\makebox[139.78bp][c]{Narrow protocol}} & \multicolumn{10}{@{}c@{}}{\includegraphics[scale=0.975,trim=0 48.665bp 0 0,clip]{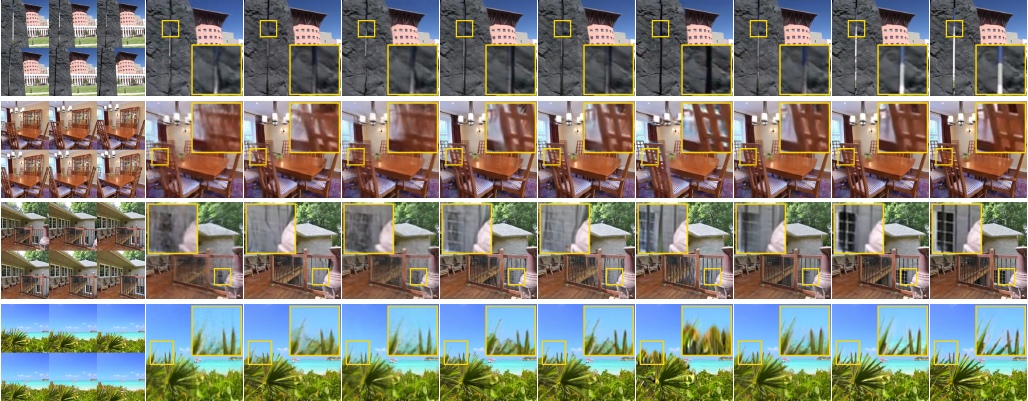}} \\[3pt]
\rotatebox{90}{\makebox[92.33bp][c]{Wide protocol}} & \multicolumn{10}{@{}c@{}}{\vbox{\offinterlineskip\hbox{\includegraphics[scale=0.975]{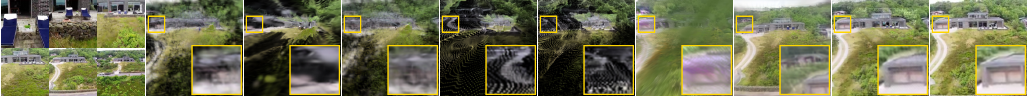}}\vskip 2.56bp\hbox{\includegraphics[scale=0.975]{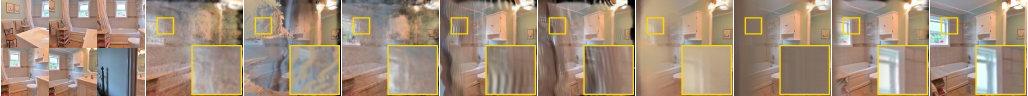}}}} \\
\end{tabular}
\caption{\textbf{Qualitative comparison on RealEstate10K.}
The upper three rows are scenes of the narrow protocol and the lower two scenes of the wide protocol.
Each row shows the six input views, one target view rendered by every method and the ground truth.
The boxed region is magnified in the opposite corner, and AnySplat, which renders from its own estimated cameras, is not shown.}
\label{fig::qual}
\end{figure*}

\subsection{In-Forward Gradient Refinement}
\label{sec::operator}

The refinement takes $K_a=\Kapp$ steps on the appearance and then $K_s=\Kshape$ steps on the shape.
Each group of steps acts on a vector $\bm{r}$ of raw network outputs, and $\theta(\bm{r})$ denotes the Gaussians that $\bm{r}$ gives with all other parameters fixed.
Starting from the predicted $\bm{r}_0$, step $k$ of a group renders $\theta(\bm{r}_k)$ into the input views and backpropagates $\mathcal{L}_{\mathrm{in}}$ through the rasterizer,
\begin{equation}
\label{eq::ctx}
\bm{g}_k=\nabla_{\bm{r}}\,\mathcal{L}_{\mathrm{in}}\bigl(\theta(\bm{r})\bigr)\Big|_{\bm{r}=\bm{r}_k}.
\end{equation}
A learned update $\mathcal{U}_\psi$ with weights $\psi$ turns the gradient into the next iterate,
\begin{equation}
\label{eq::update}
\begin{aligned}
(\Delta\bm{r}_k,\bm{h}_k)&=\mathcal{U}_\psi\bigl(\bm{r}_k,\hat{\bm{g}}_k,\Delta\bm{r}_{k-1},\bm{v}_k,\bm{F},\bm{h}_{k-1}\bigr),\\
\bm{r}_{k+1}&=\bm{r}_k+\Delta\bm{r}_k ,
\end{aligned}
\end{equation}
with $\hat{\bm{g}}_k$ the gradient divided by its root mean square over the pixels of each view and channel, $\Delta\bm{r}_{k-1}$ the previous output of $\mathcal{U}_\psi$, $\bm{v}_k=0.9\,\bm{v}_{k-1}+0.1\,\hat{\bm{g}}_k^{2}$ a running second moment with an element-wise square, and $\bm{h}_k$ a hidden state.
The recurrence starts from $\bm{h}_{-1}=\bm{0}$, $\Delta\bm{r}_{-1}=\bm{0}$ and $\bm{v}_0=\hat{\bm{g}}_0^{2}$.
The dashed box of \cref{fig::method} shows one step.
The inputs $\hat{\bm{g}}_k$, $\Delta\bm{r}_{k-1}$ and $\bm{v}_k$ are treated as constants, so training needs no second derivative of the rasterizer.
The steps read the input views only, so they run unchanged at inference.

\noindent\textbf{Learned update.} Each group has its own update of the form of \cref{eq::update}, shared by all steps of the group.
The update encodes its inputs with convolutions, refines the encoding with four transformer blocks that alternate attention within a view and across all views, and passes it to a convolutional gated recurrent unit that updates $\bm{h}_k$.
An output layer initialized to zero maps $\bm{h}_k$ to $\Delta\bm{r}_k$.
The update runs at the resolution of $\bm{F}$.

\noindent\textbf{Appearance steps.} The first group acts on the outputs of the dense head before their activations, and $\bm{r}_0$ is their predicted value.
Their gradient is taken with the base scale $\bar{\bm{s}}$ and rotation $\bar{\bm{R}}$, without the correction of \cref{eq::aniso}.
Only the opacity and color channels of $\Delta\bm{r}_k$ are applied.
Pixel unshuffling brings $\bm{r}_k$ and $\bm{g}_k$ to the resolution of $\bm{F}$, and pixel shuffling brings $\Delta\bm{r}_k$ back.

\noindent\textbf{Shape steps.} The second group acts on $\bm{r}=(\bm{r}^{\delta},\bm{r}^{\omega})$ of all Gaussians, starting from the output $\bm{r}_0$ of the shape head, with the opacity and color from the appearance steps.
Here $\theta(\bm{r})$ follows \cref{eq::aniso}, and $\bm{g}_k$ includes its Jacobian.
The steps change $\bm{r}$, not $\bm{\delta}$, so every iterate keeps $\bm{1}^{\top}\bm{\delta}=0$.
The inputs $\bm{r}_k$ and $\bm{g}_k$ are bilinearly downsampled before the normalization, and $\Delta\bm{r}_k$ is bilinearly upsampled before it is added.

\begin{table*}[t]\centering\footnotesize
\setlength{\tabcolsep}{5.2pt}
\caption{\textbf{Novel-view synthesis on RealEstate10K.} PSNR$\uparrow$, SSIM$\uparrow$ and LPIPS$\downarrow$ on the narrow and wide protocols with six input views, and PSNR on the view-count index for $V$ input views.
The upper block gives every method the input poses and the lower block gives none.
In the lower block AnySplat and \method also estimate the intrinsics and the target cameras, \method with the constant $\kappa$ shared by all scenes.
YoNoSplat$^{\dagger}$ keeps the ground-truth intrinsics and target cameras.
\colorbox{best}{\textbf{Best}} and \colorbox{second}{second best} are shaded per block, and a block of two rows shades the best only.
$\S$ marks two-view weights, $\ddagger$ a training mixture that contains DL3DV and not RealEstate10K, and $\times$ weights whose view embedding holds six views and cannot take more.}
\label{tab::main}
\begin{tabular}{lccccccccccc}
\toprule
 & \multicolumn{3}{c}{Narrow protocol} & \multicolumn{3}{c}{Wide protocol} & \multicolumn{5}{c}{PSNR$\uparrow$ for $V$ input views}\\
\cmidrule(lr){2-4}\cmidrule(lr){5-7}\cmidrule(lr){8-12}
Method & PSNR$\uparrow$ & SSIM$\uparrow$ & LPIPS$\downarrow$ & PSNR$\uparrow$ & SSIM$\uparrow$ & LPIPS$\downarrow$ & $V{=}6$ & $12$ & $24$ & $32$ & $42$\\
\midrule
pixelSplat~\cite{charatan2024pixelsplat} & 29.143 & 0.9141 & 0.1773 & 19.748 & 0.6239 & 0.4594 & 28.968 & $\times$ & $\times$ & $\times$ & $\times$\\
MVSplat~\cite{chen2024mvsplat} & 30.140 & 0.9418 & 0.0774 & 20.983 & 0.7566 & 0.2542 & 29.954 & 28.059 & 26.666 & 26.379 & 26.009\\
TranSplat~\cite{zhang2025transplat} & 30.464 & 0.9296 & 0.1454 & 19.849 & 0.6489 & 0.4265 & 30.289 & $\times$ & $\times$ & $\times$ & $\times$\\
DepthSplat~\cite{xu2025depthsplat} & \cellcolor{second}31.418 & \cellcolor{second}0.9524 & \cellcolor{second}0.0656 & 23.018 & 0.8270 & 0.1937 & \cellcolor{second}31.241 & 29.318 & 26.978 & 26.533 & 25.594\\
VolSplat~\cite{volsplat} & 31.243 & \cellcolor{second}0.9524 & \cellcolor{second}0.0656 & 23.042 & 0.8217 & 0.2069 & 31.120 & 29.285 & 26.936 & 26.676 & 25.829\\
ReSplat$^{\S}$~\cite{resplat} & 29.139 & 0.9272 & 0.0915 & 24.681 & 0.8722 & 0.1471 & 28.995 & 27.419 & 24.559 & 24.415 & 23.675\\
YoNoSplat~\cite{yonosplat} & 30.500 & 0.9390 & 0.0729 & \cellcolor{second}25.707 & \cellcolor{second}0.8779 & \cellcolor{best}\textbf{0.1295} & 30.544 & \cellcolor{second}31.038 & \cellcolor{second}31.233 & \cellcolor{second}31.199 & \cellcolor{second}31.183\\
\textbf{\method} & \cellcolor{best}\textbf{\mainNarrowOurs} & \cellcolor{best}\textbf{0.9593} & \cellcolor{best}\textbf{0.0640} & \cellcolor{best}\textbf{\mainWideOurs} & \cellcolor{best}\textbf{0.8993} & \cellcolor{second}0.1301 & \cellcolor{best}\textbf{33.636} & \cellcolor{best}\textbf{34.631} & \cellcolor{best}\textbf{34.401} & \cellcolor{best}\textbf{34.382} & \cellcolor{best}\textbf{34.801}\\
\midrule
\multicolumn{12}{l}{Without input poses}\\
AnySplat$^{\ddagger}$~\cite{anysplat} & \mainNarrowAny & 0.6160 & 0.2949 & \mainWideAny & 0.4250 & 0.4414 & 19.523 & 20.226 & 20.397 & 20.500 & 20.592\\
YoNoSplat$^{\dagger}$~\cite{yonosplat} & \cellcolor{best}\textbf{27.816} & \cellcolor{best}\textbf{0.8982} & \cellcolor{best}\textbf{0.0848} & \cellcolor{best}\textbf{21.861} & \cellcolor{best}\textbf{0.7310} & \cellcolor{best}\textbf{0.1868} & \cellcolor{best}\textbf{27.921} & \cellcolor{second}28.063 & \cellcolor{second}28.046 & \cellcolor{second}27.832 & \cellcolor{second}27.848\\
\textbf{\method} & \cellcolor{second}26.914 & \cellcolor{second}0.8863 & \cellcolor{second}0.1053 & \cellcolor{second}20.901 & \cellcolor{second}0.6945 & \cellcolor{second}0.2426 & \cellcolor{second}26.792 & \cellcolor{best}\textbf{28.896} & \cellcolor{best}\textbf{30.098} & \cellcolor{best}\textbf{30.517} & \cellcolor{best}\textbf{31.007}\\
\bottomrule
\end{tabular}
\end{table*}

\subsection{Training}
\label{sec::loss}

Training unrolls all steps and backpropagates through them, so both updates are trained jointly with the model.
Each sample has $V$ input views, with $V$ drawn uniformly from $\{2,\dots,12\}$, and four target views.
It keeps the given cameras with probability at least $0.9$ and uses the predicted ones otherwise.
Let $\theta_m$ be the Gaussians after $m$ of the $M=K_a+K_s$ steps, so $\theta_0$ is the prediction and $\theta_M$ the output.
A set $\theta$ is rendered into the target views and compared with them by $\mathcal{L}_{\mathrm{rgb}}(\theta)=\mathcal{L}_{\mathrm{MSE}}+\lambda_{\mathrm{p}}\mathcal{L}_{\mathrm{LPIPS}}+\lambda_{\mathrm{d}}\mathcal{L}_{\mathrm{D\text{-}SSIM}}$, the mean squared error, the perceptual distance~\cite{zhang2018lpips} and the structural dissimilarity $(1-\mathrm{SSIM})/2$~\cite{wang2004ssim}.
We set $\lambda_{\mathrm{p}}=0.05$, and $\lambda_{\mathrm{d}}=0.8$ for the last $24{,}000$ training iterations and $0$ before.
Every iterate is supervised,
\begin{equation}
\label{eq::deepsup}
\mathcal{L}_{\mathrm{ds}}=\sum_{m=0}^{M}\gamma^{M-m}\,\mathcal{L}_{\mathrm{rgb}}(\theta_m),
\end{equation}
with $\gamma=0.8$, so the weight grows toward the output.

A frozen copy of VGGT-$\Omega$, run on the same input views, supervises the predicted depth and cameras.
Pose and intrinsics losses supervise the camera head against the given cameras with weights $0.1$ and $0.5$.
An opacity regularizer on $\theta_M$ with weight $0.01$ and an edge-aware depth smoothness term with weight $0.05$ are added.

\section{Experiments}
\label{sec::experiments}

\subsection{Setup}
\label{sec::setup}

We train \method for $178{,}000$ steps in four stages, the first three on $16$ GH200 GPUs and the last on four, over $50.5$ hours, on the training split of RealEstate10K~\cite{zhou2018stereo}.
Evaluation uses two published indices of RealEstate10K with six input views and eight target views per scene.
The narrow protocol uses the index released with VolSplat~\cite{volsplat}, with $7{,}128$ scenes and a median of $49$ frames between the first and the last input.
The wide protocol uses the index released with YoNoSplat~\cite{yonosplat}, with $1{,}600$ scenes and a median of $278$ frames.
All scenes of both indices are evaluated at $224\times224$ pixels.
The view-count index keeps the targets of the narrow protocol and adds input views up to $V=42$, on the $6{,}444$ scenes whose frame window holds that many.
Transfer is measured on $139$ of the $140$ scenes of the DL3DV~\cite{ling2024dl3dv} index released with ReSplat~\cite{resplat}, with six input views and eight target views per scene.
On ScanNet++~\cite{yeshwanth2023scannetpp} it is measured on the validation set, with six input views and three target views per scene.
We report PSNR, SSIM~\cite{wang2004ssim} and LPIPS~\cite{zhang2018lpips}.

Every method is re-measured under one setting with released weights and its own near and far planes.
pixelSplat, MVSplat, TranSplat and DepthSplat use the weights that the VolSplat authors release for these baselines and describe as trained with the same input views as VolSplat, which trains with six.
ReSplat uses its two-view RealEstate10K weights, and YoNoSplat its RealEstate10K weights with the input poses.
AnySplat takes images alone, and its training mixture contains DL3DV but not RealEstate10K.
\method receives the input poses except in the lower block of \cref{tab::main}.

\begin{table}[tb]\centering\footnotesize
\setlength{\tabcolsep}{2.6pt}
\caption{\textbf{Transfer to DL3DV and ScanNet++.} PSNR$\uparrow$, SSIM$\uparrow$ and LPIPS$\downarrow$ on the target views of DL3DV and of the validation set of ScanNet++.
The upper block applies the weights of \cref{tab::main} without fine-tuning, and the lower block the weights that each method trains on DL3DV.
YoNoSplat is not evaluated on ScanNet++, since its input views rendered back score $13.733$\dB there, and our setting requires $20$\dB.
$\S$ marks two-view weights, $\ddagger$ a training mixture that contains DL3DV and ScanNet++ and not RealEstate10K, and -- a value that is not available.}
\label{tab::zeroshot}
\begin{tabular}{@{}lcccccc@{}}
\toprule
 & \multicolumn{3}{c}{DL3DV} & \multicolumn{3}{c}{ScanNet++ (zero-shot)}\\
\cmidrule(lr){2-4}\cmidrule(lr){5-7}
Method & PSNR$\uparrow$ & SSIM$\uparrow$ & LPIPS$\downarrow$ & PSNR$\uparrow$ & SSIM$\uparrow$ & LPIPS$\downarrow$\\
\midrule
pixelSplat~\cite{charatan2024pixelsplat} & 17.309 & 0.4671 & 0.5580 & 20.889 & 0.7038 & 0.4163\\
MVSplat~\cite{chen2024mvsplat} & 17.390 & 0.5362 & 0.4256 & 18.511 & 0.6804 & 0.3411\\
TranSplat~\cite{zhang2025transplat} & 16.964 & 0.4783 & 0.5388 & \cellcolor{second}21.458 & 0.7225 & 0.3919\\
DepthSplat~\cite{xu2025depthsplat} & 20.023 & 0.6740 & 0.3248 & 17.791 & 0.6803 & 0.3249\\
VolSplat~\cite{volsplat} & 19.813 & 0.6579 & 0.3425 & 18.745 & 0.7033 & 0.3295\\
ReSplat$^{\S}$~\cite{resplat} & 21.144 & 0.7492 & 0.2449 & 20.972 & \cellcolor{second}0.7523 & \cellcolor{best}\textbf{0.3074}\\
YoNoSplat~\cite{yonosplat} & \cellcolor{second}22.794 & \cellcolor{second}0.7694 & \cellcolor{second}0.2107 & -- & -- & --\\
AnySplat$^{\ddagger}$~\cite{anysplat} & 15.122 & 0.3390 & 0.4803 & 13.509 & 0.4368 & 0.5203\\
\textbf{\method} & \cellcolor{best}\textbf{\zsDlOurs} & \cellcolor{best}\textbf{0.8448} & \cellcolor{best}\textbf{0.1919} & \cellcolor{best}\textbf{21.609} & \cellcolor{best}\textbf{0.7540} & \cellcolor{second}0.3184\\
\midrule
\multicolumn{7}{@{}l}{Trained on DL3DV}\\
YoNoSplat~\cite{yonosplat} & 24.783 & 0.8466 & \cellcolor{best}\textbf{0.1469} & -- & -- & --\\
\textbf{\method} & \cellcolor{best}\textbf{26.104} & \cellcolor{best}\textbf{0.8802} & 0.1505 & -- & -- & --\\
\bottomrule
\end{tabular}
\end{table}

\subsection{Comparison with the State of the Art}
\label{sec::sota}
\begin{figure*}[t]\centering\scriptsize
\setlength{\tabcolsep}{0.5pt}\renewcommand{\arraystretch}{0.8}
\begin{tabular}{w{c}{0.0955\textwidth}w{c}{0.0955\textwidth}w{c}{0.0955\textwidth}w{c}{0.0955\textwidth}w{c}{0.0955\textwidth}w{c}{0.0955\textwidth}w{c}{0.0955\textwidth}w{c}{0.0955\textwidth}w{c}{0.0955\textwidth}w{c}{0.0955\textwidth}}
pixelSplat & MVSplat & DepthSplat & VolSplat & ReSplat & YoNoSplat & YoNoSplat$^{\star}$ & Ours & Ours$^{\star}$ & GT \\
\multicolumn{10}{@{}c@{}}{\includegraphics{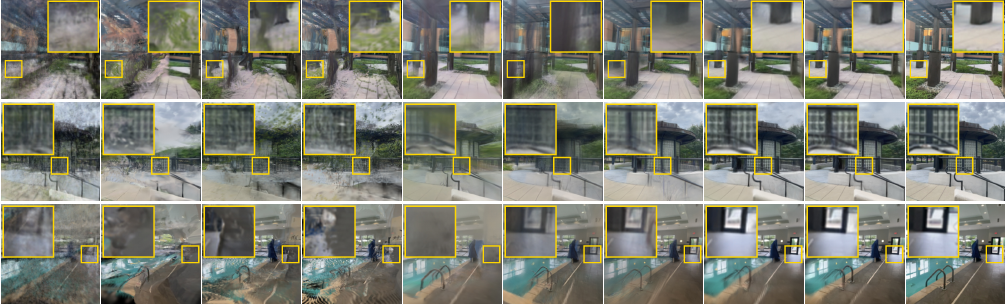}} \\
\end{tabular}
\caption{\textbf{Qualitative comparison on DL3DV.}
Each row shows one target view rendered by every method and the ground truth, and the boxed region is magnified as in \cref{fig::qual}.
$\star$ marks weights trained on DL3DV, all other columns use RealEstate10K weights, and TranSplat and AnySplat are not shown.}
\label{fig::zs-dl3dv}
\par\vspace{2.5mm}

  \centering
  \small
\setlength{\tabcolsep}{1pt}\renewcommand{\arraystretch}{0.8}
\begin{tabular}{w{c}{0.0547\textwidth}w{c}{0.082\textwidth}w{c}{0.082\textwidth}w{c}{0.082\textwidth}w{c}{0.0547\textwidth}w{c}{0.082\textwidth}w{c}{0.082\textwidth}w{c}{0.082\textwidth}w{c}{0.0547\textwidth}w{c}{0.082\textwidth}w{c}{0.082\textwidth}w{c}{0.082\textwidth}}
Inputs & AnySplat & Ours & GT & Inputs & AnySplat & Ours & GT & Inputs & AnySplat & Ours & GT \\
\multicolumn{12}{@{}c@{}}{\includegraphics{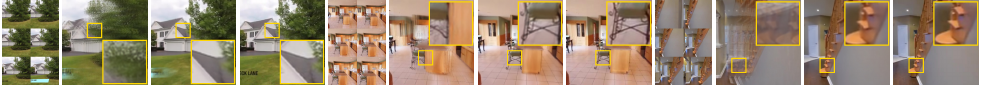}} \\
\end{tabular}
  \caption{\textbf{Renderings without input poses.} Three scenes of the narrow protocol with their input views, one target view rendered by AnySplat and by \method, and the ground truth.
  Neither method receives the input poses, and the boxed region is magnified as in \cref{fig::qual}.}
  \label{fig::posefree-qual}
\end{figure*}
\begin{figure*}[t]
  \centering
  \scriptsize
\setlength{\tabcolsep}{0.5pt}\renewcommand{\arraystretch}{0.8}
\begin{tabular}{w{c}{0.0955\textwidth}w{c}{0.0955\textwidth}w{c}{0.0955\textwidth}w{c}{0.0955\textwidth}w{c}{0.0955\textwidth}w{c}{0.0955\textwidth}w{c}{0.0955\textwidth}w{c}{0.0955\textwidth}}
pixelSplat & MVSplat & TranSplat & DepthSplat & VolSplat & ReSplat & Ours & GT \\
\multicolumn{8}{@{}c@{}}{\includegraphics{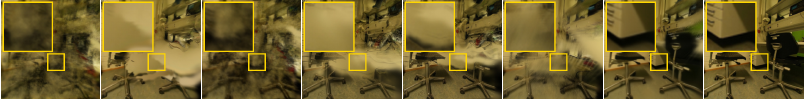}} \\
\multicolumn{8}{@{}c@{}}{\includegraphics{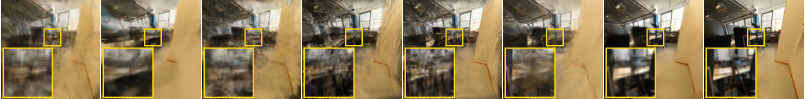}} \\
\multicolumn{8}{@{}c@{}}{\includegraphics{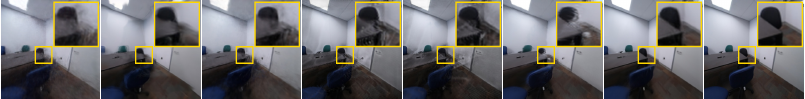}} \\
\end{tabular}
  \caption{\textbf{Qualitative comparison on ScanNet++ (zero-shot).}
  Every method uses its RealEstate10K weights without fine-tuning.
  Each row shows one target view of the validation set rendered by every method and the ground truth, and the boxed region is magnified as in \cref{fig::qual}.
  YoNoSplat and AnySplat are not shown.}
  \label{fig::zs-spp}
\end{figure*}

\noindent\textbf{Six-view protocols.} MVSplat, DepthSplat and VolSplat locate surfaces by matching the input views in a plane-sweep cost volume.
The narrow protocol of \cref{tab::main} uses the index released with VolSplat, whose inputs lie close together, and VolSplat trains with six input views that span $20$ to $50$ frames.
\method has the best PSNR, SSIM and LPIPS there, and DepthSplat and VolSplat follow it on all three.
On the wide protocol the inputs lie far apart, every method scores lower, and the order behind \method changes.
The three cost-volume methods fall behind YoNoSplat and ReSplat on every metric, whereas DepthSplat and VolSplat are ahead of both on the narrow protocol.
YoNoSplat and \method fuse the views by attention in a geometry backbone, and they are the two best methods on every metric of the wide protocol, with \method ahead on PSNR and SSIM and YoNoSplat on LPIPS.
The scenes of \cref{fig::qual} are chosen for a visible difference among those on which \method has the highest PSNR.
On the narrow protocol all methods recover the layout and differ in thin structures, such as the pole in front of the rock in the first row and the window bars behind the railing in the third.
On the wide protocol the other methods blur or break the house above the slope in the fourth row and the window of the bathroom in the fifth, which \method keeps.

\noindent\textbf{Input view count.} The right block of \cref{tab::main} adds input views to fixed targets, so a change in PSNR comes from the inputs alone.
\method has the highest PSNR at every count and stays above its six-view value at every larger one, although training draws at most $12$ views.
No parameter of \method depends on the number of views, and $\mathcal{L}_{\mathrm{in}}$ averages over them.
DepthSplat, VolSplat and MVSplat, which use the weights of the VolSplat release, lose PSNR with every added count, and so does ReSplat with its two-view weights.
At $V=6$ DepthSplat and VolSplat are the closest to \method.
YoNoSplat, trained with $2$ to $32$ views, does the same and is second from $V=12$ on.

\noindent\textbf{Transfer to DL3DV and ScanNet++.} \Cref{tab::zeroshot} applies the RealEstate10K weights to DL3DV and to the validation set of ScanNet++ without fine-tuning.
On DL3DV \method has the best PSNR, SSIM and LPIPS.
On ScanNet++ it has the best PSNR and SSIM, and ReSplat has the best LPIPS, followed by \method.
DepthSplat, VolSplat and MVSplat have a lower PSNR than \method on both datasets.
TranSplat and pixelSplat, which both run the pixelSplat encoder, are closer to \method on ScanNet++ than on DL3DV, and TranSplat is second on PSNR there.
The scenes of \cref{fig::zs-dl3dv,fig::zs-spp} are chosen in the same way as those of \cref{fig::qual}.
On DL3DV the renderings of pixelSplat, MVSplat, DepthSplat and VolSplat lose most of the layout, and those of ReSplat and YoNoSplat are blurred inside the magnified windows.
\method keeps the layout with RealEstate10K weights.
On ScanNet++ the other methods blur or double the structures inside the magnified windows, and \method keeps their edges.
The lower block compares the DL3DV weights released with YoNoSplat with \method after $12{,}000$ steps of fine-tuning on DL3DV.
\method has the higher PSNR and SSIM, and YoNoSplat the lower LPIPS.
Training on DL3DV sharpens the renderings of both methods.

\subsection{Pose-Free Inference}
\label{sec::posefree}\label{sec::posesens}

The lower block of \cref{tab::main} gives no method the input poses.
\method follows the protocol of AnySplat, where the Gaussians come from the input views alone and the camera head estimates the target cameras from the input and target images together.
It is better than AnySplat on every metric on both protocols.
YoNoSplat keeps the ground-truth intrinsics in its default pose-free mode and is rendered at the ground-truth target cameras.
It is better than \method on every metric of both six-view protocols.
\method is less accurate without input poses than with them on every metric of both protocols.
With more input views the order changes.
YoNoSplat stays close to its six-view value, while \method gains PSNR with every added count and is ahead from $V=12$ on.
\Cref{fig::posefree-qual} shows three scenes of the narrow protocol rendered without input poses.
Inside the magnified windows the renderings of AnySplat are blurred, and those of \method follow the ground truth.

\subsection{Ablation}
\label{sec::ablation}
\noindent\textbf{Training stages.} \Cref{tab::stageeval} evaluates the model after each of its four stages on the complete indices.
The first stage trains the base predictor, which has neither the shape head nor the refinement.
Every later stage raises PSNR and lowers LPIPS on both protocols.
The stage that adds the shape head and three shape steps gives the largest gain on both.
\begin{table}[tb]\centering\footnotesize
\setlength{\tabcolsep}{3.6pt}
\caption{\textbf{Training stages.} PSNR$\uparrow$ and LPIPS$\downarrow$ on the complete narrow and wide indices after each of the four training stages.
Each row continues training from the row above and adds the listed parts, the last row is \method, and Steps is the cumulative number of training steps in thousands.}
\label{tab::stageeval}
\begin{tabular}{lccccc}
\toprule
 & & \multicolumn{2}{c}{Narrow protocol} & \multicolumn{2}{c}{Wide protocol}\\
\cmidrule(lr){3-4}\cmidrule(lr){5-6}
Model & Steps & PSNR$\uparrow$ & LPIPS$\downarrow$ & PSNR$\uparrow$ & LPIPS$\downarrow$\\
\midrule
Base predictor & 100 & 28.156 & 0.0957 & 23.501 & 0.1676\\
+ shape head, $K_s{=}3$ & 130 & 30.802 & 0.0783 & 25.439 & 0.1436\\
+ $K_a{=}3$, $K_s{=}5$ & 154 & 31.844 & 0.0713 & 26.008 & 0.1360\\
+ D-SSIM, attention & 178 & \cellcolor{best}\textbf{33.683} & \cellcolor{best}\textbf{0.0640} & \cellcolor{best}\textbf{26.755} & \cellcolor{best}\textbf{0.1301}\\
\bottomrule
\end{tabular}
\end{table}

\noindent\textbf{Step count and order.} \Cref{tab::ablsteps} repeats the third training stage with other steps.
Without steps the model scores $28.583$\dB on the narrow protocol and $23.959$\dB on the wide protocol, below every row with steps.
Three shape steps score higher than three appearance steps.
In either order, PSNR rises from one step of each kind to three, and again to five shape and three appearance steps.
With the same steps, taking the appearance steps first is better than taking the shape steps first on both protocols and in both metrics.
With five shape and three appearance steps this order reaches $31.844$\dB and $26.008$\dB, the highest of the table, and \method uses it.

\noindent\textbf{Refinement steps.} We render the Gaussians of the final model after every step at inference, on the target views and on the input views, over the complete indices.
The steps read only the input views, and PSNR rises on the target views together with the input views on both protocols.
The appearance steps raise the PSNR of the target views from $29.731$ to $31.813$\dB on the narrow protocol and from $24.450$ to $25.260$\dB on the wide protocol, and the shape steps raise it to the values of \method in \cref{tab::main}.
Every shape step raises PSNR on the target and input views of both protocols, and each gains less than the one before.

\section{Conclusion}
\label{sec::conclusion}
\begin{table}[tb]\centering\footnotesize
\setlength{\tabcolsep}{4.2pt}
\caption{\textbf{Step count and order.} PSNR$\uparrow$ and LPIPS$\downarrow$ on the complete indices when the third stage of \cref{tab::stageeval} is repeated with $K_s$ shape steps and $K_a$ appearance steps, from the same weights and for the same number of training steps.
First names the kind of step that is taken first, and the last row is the third stage itself.}
\label{tab::ablsteps}
\begin{tabular}{cclcccc}
\toprule
 & & & \multicolumn{2}{c}{Narrow protocol} & \multicolumn{2}{c}{Wide protocol}\\
\cmidrule(lr){4-5}\cmidrule(lr){6-7}
$K_s$ & $K_a$ & First & PSNR$\uparrow$ & LPIPS$\downarrow$ & PSNR$\uparrow$ & LPIPS$\downarrow$\\
\midrule
0 & 0 & -- & 28.583 & 0.0927 & 23.959 & 0.1583\\
3 & 0 & -- & 31.031 & 0.0772 & 25.643 & 0.1417\\
0 & 3 & -- & 30.254 & 0.0797 & 24.797 & 0.1458\\
\midrule
\multirow{2}{*}{1} & \multirow{2}{*}{1} & shape & 30.260 & 0.0798 & 25.131 & 0.1441\\
 &  & appearance & 30.855 & 0.0778 & 25.384 & 0.1420\\
\midrule
\multirow{2}{*}{3} & \multirow{2}{*}{3} & shape & 31.202 & 0.0747 & 25.687 & 0.1398\\
 &  & appearance & \cellcolor{second}31.640 & \cellcolor{second}0.0726 & 25.837 & \cellcolor{second}0.1374\\
\midrule
\multirow{2}{*}{5} & \multirow{2}{*}{3} & shape & 31.325 & 0.0736 & \cellcolor{second}25.856 & 0.1381\\
 &  & appearance & \cellcolor{best}\textbf{31.844} & \cellcolor{best}\textbf{0.0713} & \cellcolor{best}\textbf{26.008} & \cellcolor{best}\textbf{0.1360}\\
\bottomrule
\end{tabular}
\end{table}

The Gaussians of our feed-forward predictor are less accurate than the same Gaussians optimized per scene, and their shape, the scale and rotation, carries the gap.
\method refines the prediction through two modules, log-anisotropy correction and in-forward gradient refinement, which put a learned optimizer inside the network and apply it to opacity and color and then to the shape.
The resulting feed-forward model takes learned gradient steps on the input views inside its forward pass, with no target view.
On both complete six-view RealEstate10K protocols, it has the best PSNR and SSIM among the evaluated methods and keeps its accuracy as views are added.
The refinement draws its signal from the input views and has less of it when they are few, and improving it for sparse input is the next step.

\newpage
\FloatBarrier   %
{
\small
\noindent\textbf{Acknowledgements.} This work was supported by the U.S. National Science Foundation under Grant No.~2316400.
The authors also acknowledge the South Dakota State University High Performance Computing (SDSU HPC) team for providing access to and support for the university's high-performance computing resources.
We acknowledge CSC -- IT Center for Science, Finland, for providing access to Roihu and to LUMI, which is owned by the EuroHPC Joint Undertaking and hosted by CSC -- IT Center for Science, Finland, in collaboration with the LUMI consortium.
\par
}

{
    \small
    \bibliographystyle{ieeenat_fullname}
    \bibliography{references}
}

\setcounter{dbltopnumber}{3}
\clearpage
\setcounter{figure}{0}\renewcommand{\thefigure}{A\arabic{figure}}
\setcounter{table}{0}\renewcommand{\thetable}{A\arabic{table}}
\setcounter{equation}{0}\renewcommand{\theequation}{A\arabic{equation}}
\setcounter{dbltopnumber}{2}
\newenvironment{titlefloat}[1]{\par\noindent\begin{minipage}{\textwidth}\captionsetup{type=#1}}{\end{minipage}\par}
\newcommand{\supptitletable}{\begingroup
  \renewenvironment{table*}[1][]{\begin{titlefloat}{table}}{\end{titlefloat}}%
  \begin{table*}[!tb]\centering\footnotesize
\setlength{\tabcolsep}{4.6pt}
\caption{\textbf{Complete comparison with six input views.} PSNR$\uparrow$, SSIM$\uparrow$ and LPIPS$\downarrow$ on the complete narrow and wide protocols of RealEstate10K, on $139$ scenes of DL3DV and on the validation set of ScanNet++, with the weights of \cref{tab::main} and no fine-tuning.
YoNoSplat has no ScanNet++ entry, since its self-reconstruction there is $13.733$\dB and our setting requires $20$\dB.
$\S$ marks two-view weights, $\ddagger$ a training mixture that contains DL3DV and ScanNet++ and not RealEstate10K, and -- a setting without a reported value.}
\label{tab::all}
\begin{tabular}{lcccccccccccc}
\toprule
 & \multicolumn{3}{c}{RealEstate10K narrow} & \multicolumn{3}{c}{RealEstate10K wide} & \multicolumn{3}{c}{DL3DV} & \multicolumn{3}{c}{ScanNet++}\\
\cmidrule(lr){2-4}\cmidrule(lr){5-7}\cmidrule(lr){8-10}\cmidrule(lr){11-13}
Method & PSNR$\uparrow$ & SSIM$\uparrow$ & LPIPS$\downarrow$ & PSNR$\uparrow$ & SSIM$\uparrow$ & LPIPS$\downarrow$ & PSNR$\uparrow$ & SSIM$\uparrow$ & LPIPS$\downarrow$ & PSNR$\uparrow$ & SSIM$\uparrow$ & LPIPS$\downarrow$\\
\midrule
pixelSplat & 29.143 & 0.9141 & 0.1773 & 19.748 & 0.6239 & 0.4594 & 17.309 & 0.4671 & 0.5580 & 20.889 & 0.7038 & 0.4163\\
MVSplat & 30.140 & 0.9418 & 0.0774 & 20.983 & 0.7566 & 0.2542 & 17.390 & 0.5362 & 0.4256 & 18.511 & 0.6804 & 0.3411\\
TranSplat & 30.464 & 0.9296 & 0.1454 & 19.849 & 0.6489 & 0.4265 & 16.964 & 0.4783 & 0.5388 & \cellcolor{second}21.458 & 0.7225 & 0.3919\\
DepthSplat & \cellcolor{second}31.418 & \cellcolor{second}0.9524 & \cellcolor{second}0.0656 & 23.018 & 0.8270 & 0.1937 & 20.023 & 0.6740 & 0.3248 & 17.791 & 0.6803 & 0.3249\\
VolSplat & 31.243 & \cellcolor{second}0.9524 & \cellcolor{second}0.0656 & 23.042 & 0.8217 & 0.2069 & 19.813 & 0.6579 & 0.3425 & 18.745 & 0.7033 & 0.3295\\
ReSplat$^{\S}$ & 29.139 & 0.9272 & 0.0915 & 24.681 & 0.8722 & 0.1471 & 21.144 & 0.7492 & 0.2449 & 20.972 & \cellcolor{second}0.7523 & \cellcolor{best}\textbf{0.3074}\\
YoNoSplat & 30.500 & 0.9390 & 0.0729 & \cellcolor{second}25.707 & \cellcolor{second}0.8779 & \cellcolor{best}\textbf{0.1295} & \cellcolor{second}22.794 & \cellcolor{second}0.7694 & \cellcolor{second}0.2107 & -- & -- & --\\
AnySplat$^{\ddagger}$ & \mainNarrowAny & 0.6160 & 0.2949 & \mainWideAny & 0.4250 & 0.4414 & 15.122 & 0.3390 & 0.4803 & 13.509 & 0.4368 & 0.5203\\
\textbf{\method} & \cellcolor{best}\textbf{\mainNarrowOurs} & \cellcolor{best}\textbf{0.9593} & \cellcolor{best}\textbf{0.0640} & \cellcolor{best}\textbf{\mainWideOurs} & \cellcolor{best}\textbf{0.8993} & \cellcolor{second}0.1301 & \cellcolor{best}\textbf{\zsDlOurs} & \cellcolor{best}\textbf{0.8448} & \cellcolor{best}\textbf{0.1919} & \cellcolor{best}\textbf{21.609} & \cellcolor{best}\textbf{0.7540} & \cellcolor{second}0.3184\\
\bottomrule
\end{tabular}
\end{table*}
  \endgroup}
\newcommand{\supptitlefigure}{%
  \begin{titlefloat}{figure}
    \centering
    \small
  \setlength{\tabcolsep}{0.5pt}\renewcommand{\arraystretch}{0.8}
  \begin{tabular}{w{c}{0.1395\textwidth}w{c}{0.093\textwidth}w{c}{0.093\textwidth}w{c}{0.093\textwidth}w{c}{0.093\textwidth}w{c}{0.093\textwidth}w{c}{0.093\textwidth}w{c}{0.093\textwidth}w{c}{0.093\textwidth}w{c}{0.093\textwidth}}
  Inputs & pixelSplat & MVSplat & TranSplat & DepthSplat & VolSplat & ReSplat & YoNoSplat & Ours & GT \\
  \multicolumn{10}{@{}c@{}}{\includegraphics{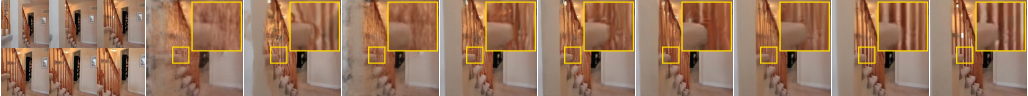}} \\
   & 23.12 & 23.61 & 24.39 & 24.13 & 24.50 & 23.64 & 25.82 & 28.91 &  \\
  \multicolumn{10}{@{}c@{}}{\includegraphics{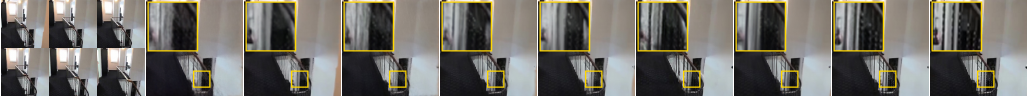}} \\
   & 26.90 & 25.33 & 27.43 & 27.01 & 26.55 & 24.10 & 27.66 & 31.73 &  \\
  \multicolumn{10}{@{}c@{}}{\includegraphics{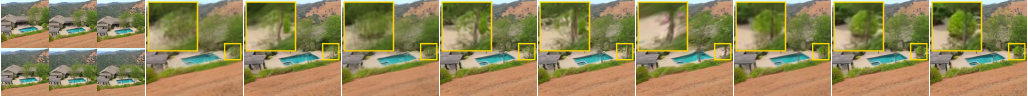}} \\
   & 25.98 & 27.92 & 26.60 & 28.27 & 27.78 & 24.27 & 28.06 & 32.10 &  \\
  \end{tabular}
    \caption{\textbf{Further scenes of the narrow protocol.} Each row shows the six input views, one target view rendered by every method and the ground truth, as in \cref{fig::qual}, with the PSNR of the scene beneath each rendering.}
    \label{fig::qual-supp}
  \end{titlefloat}}
\twocolumn[%
  {\centering \Large \textbf{\thetitle}\\ \vspace{0.5em}Supplementary Material \\ \vspace{1.0em}\par}
  \begingroup
  \renewenvironment{table*}[1][]{\begin{titlefloat}{table}}{\end{titlefloat}}%
  \begin{table*}[!tb]\centering\footnotesize
\setlength{\tabcolsep}{4.6pt}
\caption{\textbf{Complete comparison with six input views.} PSNR$\uparrow$, SSIM$\uparrow$ and LPIPS$\downarrow$ on the complete narrow and wide protocols of RealEstate10K, on $139$ scenes of DL3DV and on the validation set of ScanNet++, with the weights of \cref{tab::main} and no fine-tuning.
YoNoSplat has no ScanNet++ entry, since its self-reconstruction there is $13.733$\dB and our setting requires $20$\dB.
$\S$ marks two-view weights, $\ddagger$ a training mixture that contains DL3DV and ScanNet++ and not RealEstate10K, and -- a setting without a reported value.}
\label{tab::all}
\begin{tabular}{lcccccccccccc}
\toprule
 & \multicolumn{3}{c}{RealEstate10K narrow} & \multicolumn{3}{c}{RealEstate10K wide} & \multicolumn{3}{c}{DL3DV} & \multicolumn{3}{c}{ScanNet++}\\
\cmidrule(lr){2-4}\cmidrule(lr){5-7}\cmidrule(lr){8-10}\cmidrule(lr){11-13}
Method & PSNR$\uparrow$ & SSIM$\uparrow$ & LPIPS$\downarrow$ & PSNR$\uparrow$ & SSIM$\uparrow$ & LPIPS$\downarrow$ & PSNR$\uparrow$ & SSIM$\uparrow$ & LPIPS$\downarrow$ & PSNR$\uparrow$ & SSIM$\uparrow$ & LPIPS$\downarrow$\\
\midrule
pixelSplat & 29.143 & 0.9141 & 0.1773 & 19.748 & 0.6239 & 0.4594 & 17.309 & 0.4671 & 0.5580 & 20.889 & 0.7038 & 0.4163\\
MVSplat & 30.140 & 0.9418 & 0.0774 & 20.983 & 0.7566 & 0.2542 & 17.390 & 0.5362 & 0.4256 & 18.511 & 0.6804 & 0.3411\\
TranSplat & 30.464 & 0.9296 & 0.1454 & 19.849 & 0.6489 & 0.4265 & 16.964 & 0.4783 & 0.5388 & \cellcolor{second}21.458 & 0.7225 & 0.3919\\
DepthSplat & \cellcolor{second}31.418 & \cellcolor{second}0.9524 & \cellcolor{second}0.0656 & 23.018 & 0.8270 & 0.1937 & 20.023 & 0.6740 & 0.3248 & 17.791 & 0.6803 & 0.3249\\
VolSplat & 31.243 & \cellcolor{second}0.9524 & \cellcolor{second}0.0656 & 23.042 & 0.8217 & 0.2069 & 19.813 & 0.6579 & 0.3425 & 18.745 & 0.7033 & 0.3295\\
ReSplat$^{\S}$ & 29.139 & 0.9272 & 0.0915 & 24.681 & 0.8722 & 0.1471 & 21.144 & 0.7492 & 0.2449 & 20.972 & \cellcolor{second}0.7523 & \cellcolor{best}\textbf{0.3074}\\
YoNoSplat & 30.500 & 0.9390 & 0.0729 & \cellcolor{second}25.707 & \cellcolor{second}0.8779 & \cellcolor{best}\textbf{0.1295} & \cellcolor{second}22.794 & \cellcolor{second}0.7694 & \cellcolor{second}0.2107 & -- & -- & --\\
AnySplat$^{\ddagger}$ & \mainNarrowAny & 0.6160 & 0.2949 & \mainWideAny & 0.4250 & 0.4414 & 15.122 & 0.3390 & 0.4803 & 13.509 & 0.4368 & 0.5203\\
\textbf{\method} & \cellcolor{best}\textbf{\mainNarrowOurs} & \cellcolor{best}\textbf{0.9593} & \cellcolor{best}\textbf{0.0640} & \cellcolor{best}\textbf{\mainWideOurs} & \cellcolor{best}\textbf{0.8993} & \cellcolor{second}0.1301 & \cellcolor{best}\textbf{\zsDlOurs} & \cellcolor{best}\textbf{0.8448} & \cellcolor{best}\textbf{0.1919} & \cellcolor{best}\textbf{21.609} & \cellcolor{best}\textbf{0.7540} & \cellcolor{second}0.3184\\
\bottomrule
\end{tabular}
\end{table*}
  \endgroup
  \vspace{\dblfloatsep}
  \supptitlefigure
  \vspace{2pt}
]
\appendix

This supplement first gives further comparisons, with all six-view results in one table, more scenes of both RealEstate10K protocols, ScanNet++~\cite{yeshwanth2023scannetpp}, two input views, more input views and inference without input poses.
It then reports the measurements behind the design of the refinement, the implementation details of \method, and the evaluation protocols and baseline weights.

\section{Additional Comparisons}
\label{sec::supp-compare}

\begin{figure*}[t]
  \centering
  \small
\setlength{\tabcolsep}{0.5pt}\renewcommand{\arraystretch}{0.8}
\begin{tabular}{w{c}{0.1395\textwidth}w{c}{0.093\textwidth}w{c}{0.093\textwidth}w{c}{0.093\textwidth}w{c}{0.093\textwidth}w{c}{0.093\textwidth}w{c}{0.093\textwidth}w{c}{0.093\textwidth}w{c}{0.093\textwidth}w{c}{0.093\textwidth}}
Inputs & pixelSplat & MVSplat & TranSplat & DepthSplat & VolSplat & ReSplat & YoNoSplat & Ours & GT \\
\multicolumn{10}{@{}c@{}}{\includegraphics{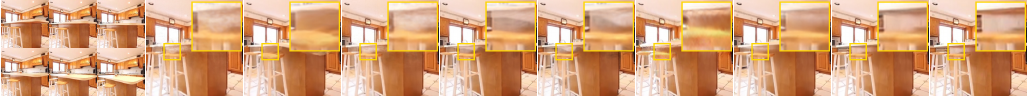}} \\
 & 27.79 & 25.15 & 29.13 & 28.77 & 29.53 & 24.16 & 29.77 & 33.50 &  \\
\multicolumn{10}{@{}c@{}}{\includegraphics{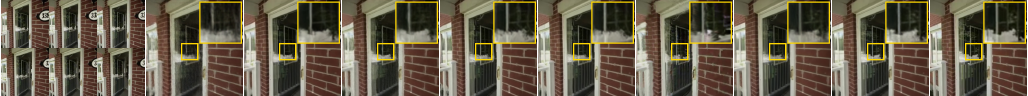}} \\
 & 27.23 & 28.54 & 29.22 & 30.07 & 29.62 & 23.66 & 26.31 & 33.53 &  \\
\multicolumn{10}{@{}c@{}}{\includegraphics{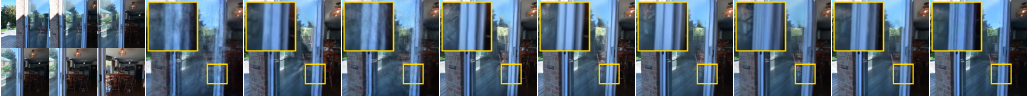}} \\
 & 22.31 & 21.43 & 23.32 & 22.87 & 22.77 & 23.77 & 23.69 & 27.17 &  \\
\end{tabular}
  \caption{\textbf{Further scenes of the narrow protocol, continued.} Layout as in \cref{fig::qual-supp}.}
  \label{fig::qual-supp2}
\end{figure*}

\begin{figure*}[t]
  \centering
  \small
\setlength{\tabcolsep}{0.5pt}\renewcommand{\arraystretch}{0.8}
\begin{tabular}{w{c}{0.1395\textwidth}w{c}{0.093\textwidth}w{c}{0.093\textwidth}w{c}{0.093\textwidth}w{c}{0.093\textwidth}w{c}{0.093\textwidth}w{c}{0.093\textwidth}w{c}{0.093\textwidth}w{c}{0.093\textwidth}w{c}{0.093\textwidth}}
Inputs & pixelSplat & MVSplat & TranSplat & DepthSplat & VolSplat & ReSplat & YoNoSplat & Ours & GT \\
\multicolumn{10}{@{}c@{}}{\includegraphics{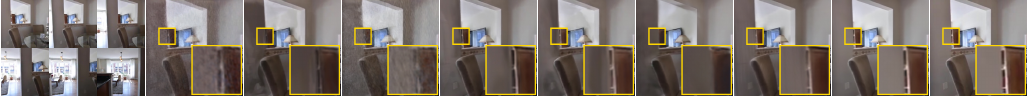}} \\
 & 23.87 & 22.20 & 24.53 & 23.04 & 22.97 & 21.21 & 25.81 & 30.81 &  \\
\multicolumn{10}{@{}c@{}}{\includegraphics{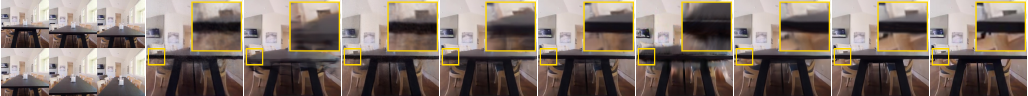}} \\
 & 29.11 & 24.55 & 30.12 & 27.95 & 30.81 & 23.72 & 30.78 & 35.63 &  \\
\multicolumn{10}{@{}c@{}}{\includegraphics{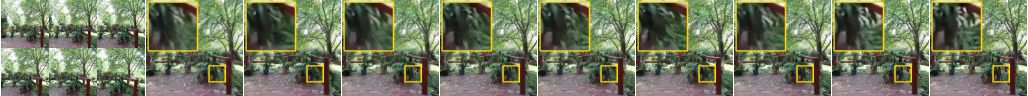}} \\
 & 24.87 & 25.20 & 25.84 & 26.54 & 26.40 & 23.51 & 25.75 & 29.14 &  \\
\end{tabular}
  \caption{\textbf{Further scenes of the narrow protocol, continued.} Layout as in \cref{fig::qual-supp}.}
  \label{fig::qual-supp3}
\end{figure*}

\begin{figure*}[t]
  \centering
  \small
\setlength{\tabcolsep}{0.5pt}\renewcommand{\arraystretch}{0.8}
\begin{tabular}{w{c}{0.1395\textwidth}w{c}{0.093\textwidth}w{c}{0.093\textwidth}w{c}{0.093\textwidth}w{c}{0.093\textwidth}w{c}{0.093\textwidth}w{c}{0.093\textwidth}w{c}{0.093\textwidth}w{c}{0.093\textwidth}w{c}{0.093\textwidth}}
Inputs & pixelSplat & MVSplat & TranSplat & DepthSplat & VolSplat & ReSplat & YoNoSplat & Ours & GT \\
\multicolumn{10}{@{}c@{}}{\includegraphics{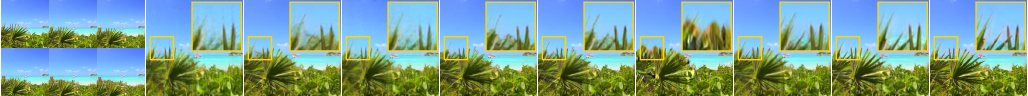}} \\
 & 24.65 & 24.89 & 25.62 & 25.68 & 25.70 & 20.42 & 25.09 & 27.69 &  \\
\multicolumn{10}{@{}c@{}}{\includegraphics{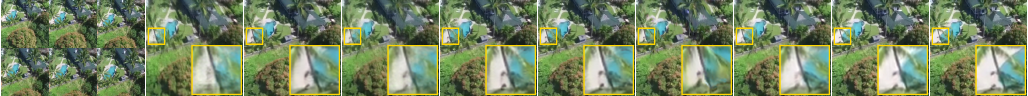}} \\
 & 24.92 & 27.16 & 25.99 & 28.03 & 28.21 & 24.03 & 27.19 & 30.94 &  \\
\multicolumn{10}{@{}c@{}}{\includegraphics{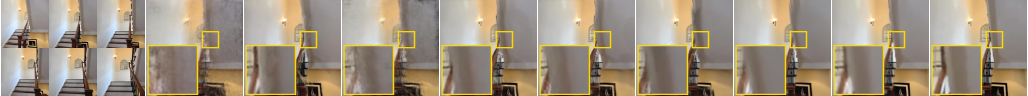}} \\
 & 23.58 & 24.28 & 24.10 & 24.09 & 23.90 & 22.61 & 23.73 & 26.59 &  \\
\end{tabular}
  \caption{\textbf{Qualitative comparison on the narrow protocol, continued.} Layout as in \cref{fig::qual-supp}.}
  \label{fig::qual-supp4}
\end{figure*}

\begin{figure*}[t]
  \centering
  \small
\setlength{\tabcolsep}{0.5pt}\renewcommand{\arraystretch}{0.8}
\begin{tabular}{w{c}{0.1395\textwidth}w{c}{0.093\textwidth}w{c}{0.093\textwidth}w{c}{0.093\textwidth}w{c}{0.093\textwidth}w{c}{0.093\textwidth}w{c}{0.093\textwidth}w{c}{0.093\textwidth}w{c}{0.093\textwidth}w{c}{0.093\textwidth}}
Inputs & pixelSplat & MVSplat & TranSplat & DepthSplat & VolSplat & ReSplat & YoNoSplat & Ours & GT \\
\multicolumn{10}{@{}c@{}}{\includegraphics{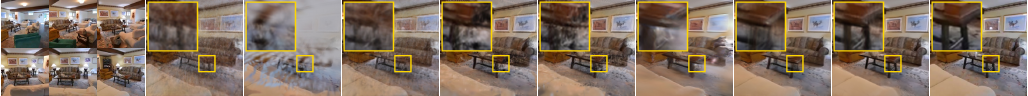}} \\
 & 19.37 & 13.57 & 20.70 & 22.38 & 22.73 & 19.68 & 24.46 & 26.97 &  \\
\multicolumn{10}{@{}c@{}}{\includegraphics{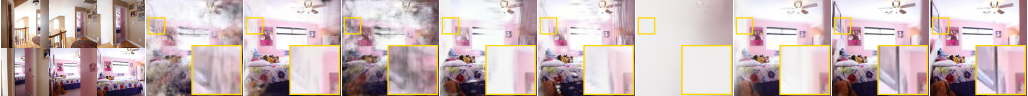}} \\
 & 16.63 & 19.23 & 16.07 & 20.15 & 20.21 & 20.03 & 23.43 & 25.81 &  \\
\multicolumn{10}{@{}c@{}}{\includegraphics{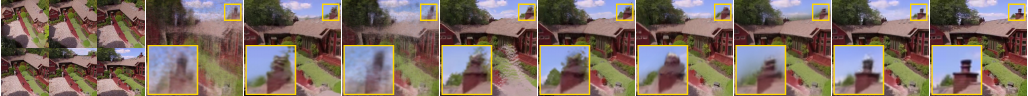}} \\
 & 20.56 & 19.30 & 21.68 & 25.09 & 25.84 & 26.02 & 26.25 & 29.57 &  \\
\end{tabular}
  \caption{\textbf{Qualitative comparison on the wide protocol.} Layout as in \cref{fig::qual}, with the PSNR of the scene beneath each rendering.}
  \label{fig::qual-wide}
\end{figure*}

\begin{figure*}[t]
  \centering
  \small
\setlength{\tabcolsep}{0.5pt}\renewcommand{\arraystretch}{0.8}
\begin{tabular}{w{c}{0.1395\textwidth}w{c}{0.093\textwidth}w{c}{0.093\textwidth}w{c}{0.093\textwidth}w{c}{0.093\textwidth}w{c}{0.093\textwidth}w{c}{0.093\textwidth}w{c}{0.093\textwidth}w{c}{0.093\textwidth}w{c}{0.093\textwidth}}
Inputs & pixelSplat & MVSplat & TranSplat & DepthSplat & VolSplat & ReSplat & YoNoSplat & Ours & GT \\
\multicolumn{10}{@{}c@{}}{\includegraphics{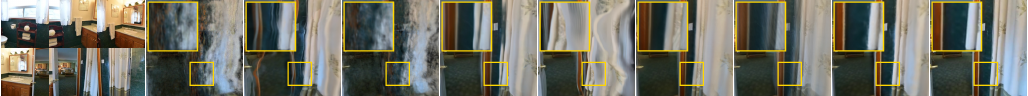}} \\
 & 18.86 & 18.03 & 19.87 & 22.75 & 19.73 & 24.17 & 24.10 & 27.44 &  \\
\multicolumn{10}{@{}c@{}}{\includegraphics{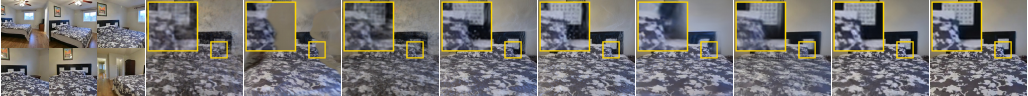}} \\
 & 16.41 & 16.18 & 16.52 & 20.01 & 19.78 & 19.94 & 20.37 & 24.78 &  \\
\multicolumn{10}{@{}c@{}}{\includegraphics{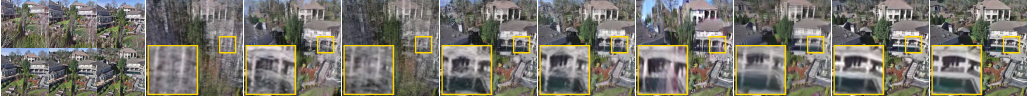}} \\
 & 17.34 & 19.04 & 17.02 & 20.27 & 19.39 & 19.17 & 20.22 & 24.72 &  \\
\end{tabular}
  \caption{\textbf{Qualitative comparison on the wide protocol, continued.} Layout as in \cref{fig::qual-wide}.}
  \label{fig::qual-wide2}
\end{figure*}

\begin{figure*}[t]
  \centering
  \small
\setlength{\tabcolsep}{1pt}\renewcommand{\arraystretch}{0.8}
\begin{tabular}{lw{c}{0.093\textwidth}w{c}{0.093\textwidth}w{c}{0.093\textwidth}w{c}{0.093\textwidth}w{c}{0.093\textwidth}w{c}{0.093\textwidth}w{c}{0.093\textwidth}w{c}{0.093\textwidth}w{c}{0.093\textwidth}w{c}{0.093\textwidth}}
 & $V=6$ & $V=12$ & $V=24$ & $V=42$ & GT & $V=6$ & $V=12$ & $V=24$ & $V=42$ & GT \\
\rowname{DepthSplat} & \multicolumn{9}{@{}c@{}}{\includegraphics{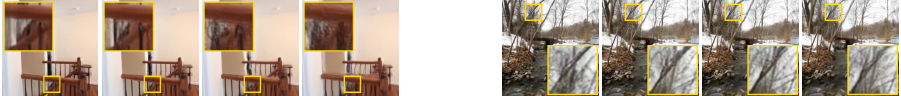}} &  \\
 & 30.82 & 28.74 & 25.44 & 24.72 &  & 27.43 & 24.90 & 22.38 & 21.97 &  \\
\rowname{YoNoSplat} & \multicolumn{9}{@{}c@{}}{\includegraphics{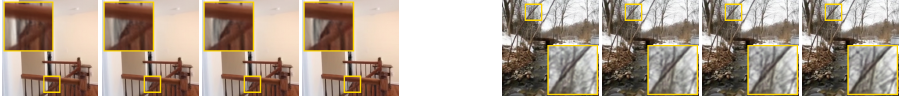}} &  \\
 & 30.83 & 31.46 & 31.81 & 31.73 &  & 26.60 & 27.11 & 27.42 & 27.54 &  \\
\rowname{Ours} & \multicolumn{10}{@{}c@{}}{\includegraphics{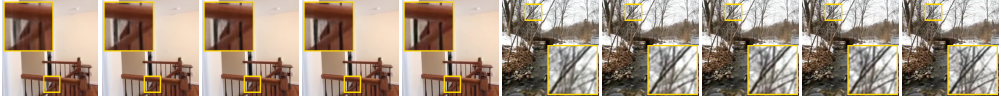}} \\
 & 35.40 & 36.25 & 36.26 & 36.94 &  & 29.77 & 31.57 & 31.96 & 32.19 &  \\
\end{tabular}
  \caption{\textbf{Renderings for an increasing number of input views.} One target view of two scenes of the view-count index, rendered by DepthSplat, YoNoSplat and \method from $V$ input views, with the PSNR of the scene beneath each rendering.
  The target is the same in every column of a scene, and a larger $V$ only adds input views.}
  \label{fig::viewcount-qual}
\end{figure*}

\paragraph{Complete comparison.}\label{sec::supp-all} \Cref{tab::all} collects the six-view results of \cref{tab::main,tab::zeroshot}, all with the weights of \cref{tab::main}.
\method has the best PSNR and SSIM on both RealEstate10K protocols and the best value of every metric on DL3DV.
On ScanNet++ it has the best PSNR and SSIM and the second-best LPIPS.
The order of the other methods differs between the two transfer datasets.
By PSNR, TranSplat and pixelSplat are eighth and seventh of nine methods on DL3DV, and second and fourth of eight on ScanNet++.

\paragraph{Qualitative figures.} The scenes of the qualitative figures of this supplement are chosen for a visible difference.
Each row shows one target view of a scene, and its magnified box is the same window for every rendering of the row.
In the comparison figures, the window lies where the gap in LPIPS between \method and the best other method is large, and the number beneath a rendering is the PSNR of the scene over all its target views.

\paragraph{Narrow protocol.} \Cref{fig::qual-supp,fig::qual-supp2,fig::qual-supp3,fig::qual-supp4} show twelve further scenes of the narrow protocol in the layout of \cref{fig::qual}.
Like the three narrow scenes of \cref{fig::qual}, they are chosen among $249$ scenes of the narrow index on which \method has the highest PSNR and the lowest LPIPS of the eight methods.

\paragraph{Wide protocol.}\label{sec::supp-wide-qual} \Cref{fig::qual-wide,fig::qual-wide2} show six further scenes of the wide protocol, on which every method of \cref{tab::main} scores lower.
Like the two wide scenes of \cref{fig::qual}, they are chosen among $160$ scenes of the wide index on which \method has the highest PSNR and the lowest LPIPS of the eight methods.

\paragraph{Training on DL3DV.} The lower block of \cref{tab::zeroshot} fine-tunes the final model on DL3DV for $12{,}000$ training steps, which takes $9.7$ hours on four GH200 GPUs.
The fine-tuning starts from the weights of the final model with a new optimizer, a learning rate of $8\times10^{-5}$, six input views and a D-SSIM weight of $0.8$.
It draws the frame span from the first of the three bands of the training sampler only.
On the $139$ scenes the PSNR is $25.726$, $25.822$, $26.135$ and $26.104$\dB after $3{,}000$, $6{,}000$, $9{,}000$ and $12{,}000$ training steps, and \cref{tab::zeroshot} reports the last of them.
After the fine-tuning the model scores $26.157$\dB PSNR, $0.8915$ SSIM and $0.1376$ LPIPS on the $1{,}600$ scenes of the wide protocol.
This PSNR is below that of the final model in \cref{tab::main}.

\paragraph{ScanNet++.}\label{sec::supp-spp} YoNoSplat has no ScanNet++ values in \cref{tab::zeroshot}, since its self-reconstruction on these scenes is $13.733$\dB, below the $20$\dB that our setting requires.
It is $23.579$\dB on DL3DV.

\begin{table}[!tb]\centering\footnotesize
\setlength{\tabcolsep}{6pt}
\caption{\textbf{Two input views.} PSNR$\uparrow$ on the complete two-view indices of RealEstate10K, ACID and DTU, with RealEstate10K weights and no fine-tuning.
ReSplat and \method are evaluated at $256\times256$ pixels and YoNoSplat and AnySplat at $224\times224$.
At $256\times256$ pixels the position embeddings of the transformer blocks of \method are interpolated bicubically from $14\times14$ to $16\times16$ tokens.
$\S$ marks two-view weights and $\ddagger$ a training mixture that contains DL3DV and ScanNet++ and not RealEstate10K.}
\label{tab::twoview}
\begin{tabular}{lccc}
\toprule
Method & RealEstate10K & ACID & DTU\\
\midrule
ReSplat$^{\S}$~\cite{resplat} & \cellcolor{best}\textbf{30.613} & \cellcolor{best}\textbf{31.285} & \cellcolor{best}\textbf{13.958}\\
YoNoSplat~\cite{yonosplat} & 26.574 & 26.783 & \cellcolor{second}12.572\\
AnySplat$^{\ddagger}$~\cite{anysplat} & 16.579 & 19.924 & 11.457\\
\textbf{\method} & \cellcolor{second}27.301 & \cellcolor{second}27.898 & 10.754\\
\bottomrule
\end{tabular}
\end{table}

\paragraph{Two-view regime.}\label{sec::supp-twoview} The refinement takes its gradient from the input views, and the protocols above give it six or more.
\Cref{tab::twoview} evaluates two input views on the two-view indices of RealEstate10K, ACID~\cite{liu2021infinite} and DTU~\cite{jensen2014dtu}, with RealEstate10K weights and no fine-tuning.
ReSplat, whose RealEstate10K weights are trained with two views, has the best PSNR on all three.
\method is below it and above YoNoSplat on RealEstate10K and ACID.
On DTU all four methods score far below their RealEstate10K values, and \method has the lowest PSNR.
DTU is object-centric, with a longer focal length than any RealEstate10K scene.
With six input views the order of ReSplat and \method is reversed in \cref{tab::main,tab::zeroshot}.
Our setting measures ReSplat above its printed two-view values in \cref{tab::calib}.

\paragraph{Input view count.}\label{sec::supp-views} From $V=6$ to $V=42$ on the view-count index, the PSNR of DepthSplat, VolSplat and MVSplat in \cref{tab::main} falls with every increase of $V$.
That of \method is the best at every count, and that of \method and that of YoNoSplat are higher at every larger count than at $V=6$.
The margin of \method over DepthSplat therefore grows with $V$, and most of this growth is the decline of DepthSplat.
The margin over YoNoSplat is larger at $V=42$ than at $V=6$.
MVSplat averages the correlations of a view with all other input views.
DepthSplat and VolSplat match a view against its two nearest views once more than three are given.
YoNoSplat alternates attention inside a view with attention over all views.
VolSplat trains with six input views, and its release describes the baseline weights as trained with the same input views.
YoNoSplat trains with $2$ to $32$ input views and \method with $2$ to $12$.
No experiment isolates which of these differences causes the decline.
pixelSplat and TranSplat cannot be evaluated beyond six views, since the view embedding in both sets of weights has six entries.
ReSplat, whose RealEstate10K weights are trained with two views, falls with every increase of $V$.
Without input poses, AnySplat and \method gain with every increase of $V$ and YoNoSplat stays close to its value at $V=6$.
Below six views \method scores $28.492$, $31.490$ and $32.501$\dB at $V=2$, $3$ and $4$ on the same $6{,}444$ scenes.
\Cref{fig::viewcount-qual} shows two scenes.
In both scenes of \cref{fig::viewcount-qual}, the PSNR of DepthSplat at $V=42$ is well below its value at $V=6$.
That of YoNoSplat and that of \method are higher at $V=42$ than at $V=6$ in both scenes.

\begin{table}[!tb]\centering\footnotesize
\setlength{\tabcolsep}{1.85pt}
\caption{\textbf{Pose-free settings.} PSNR$\uparrow$, SSIM$\uparrow$ and LPIPS$\downarrow$ on the complete narrow and wide protocols without input poses.
The upper block repeats the lower block of \cref{tab::main}, and the lower block is the final model with $\kappa=1$ and with ground-truth target cameras and aligned scale.
$\dagger$ marks ground-truth intrinsics and target cameras and $\ddagger$ a training mixture that contains DL3DV and ScanNet++ and not RealEstate10K.}
\label{tab::posefree}
\begin{tabular}{@{}lcccccc@{}}
\toprule
 & \multicolumn{3}{c}{Narrow protocol} & \multicolumn{3}{c}{Wide protocol}\\
\cmidrule(lr){2-4}\cmidrule(lr){5-7}
Setting & PSNR$\uparrow$ & SSIM$\uparrow$ & LPIPS$\downarrow$ & PSNR$\uparrow$ & SSIM$\uparrow$ & LPIPS$\downarrow$\\
\midrule
AnySplat$^{\ddagger}$ & \mainNarrowAny & 0.6160 & 0.2949 & \mainWideAny & 0.4250 & 0.4414\\
YoNoSplat$^{\dagger}$ & 27.816 & 0.8982 & 0.0848 & 21.861 & 0.7310 & 0.1868\\
\method, $\kappa{=}4.04$ & 26.914 & 0.8863 & 0.1053 & 20.901 & 0.6945 & 0.2426\\
\midrule
\method, $\kappa{=}1$ & 26.538 & 0.8724 & 0.1711 & 20.669 & 0.6638 & 0.3805\\
\method, GT targets & 19.794 & 0.6501 & 0.2345 & 17.782 & 0.5524 & 0.3193\\
\bottomrule
\end{tabular}
\end{table}

\paragraph{Pose-free protocol.}\label{sec::supp-posefree} \Cref{tab::posefree} repeats the lower block of \cref{tab::main} and adds two settings of the final model.
In the protocol of \cref{tab::main}, the Gaussians of \method come from the input views alone and are placed with the predicted cameras.
The pose and intrinsics of the target cameras come from a second pass of the backbone and the camera head over the input and target images together, as in the evaluation of AnySplat.
The target images enter the estimate of their own cameras and never the Gaussians.
No ground-truth pose is used.
The constant $\kappa$ multiplies the predicted depth and the translations of the input and target cameras.
YoNoSplat runs in the default of its two pose-free modes, which predicts the input poses, keeps the ground-truth intrinsics and renders at the ground-truth target cameras in its normalized frame.
The three rows of the upper block do not receive the same information.

\paragraph{Scale constant.} The scale of a Gaussian is expressed in the fixed unit $\tau$, while the predicted cameras fix the scene up to a global scale.
The constant $\kappa$ therefore sets the size of a Gaussian relative to the scene.
We set $\kappa=4.04$.
This is the median ratio between the ground-truth and the predicted distance of the first two input cameras, over $60$ scenes of the narrow protocol with the final model.
The constant uses a statistic of the ground-truth cameras of evaluation scenes, once for the dataset, and the same value is applied on the wide protocol.
With $\kappa=1$ on the complete indices, every metric of \cref{tab::posefree} is worse than with $\kappa=4.04$.

\paragraph{Ground-truth target cameras and intrinsics.} The last row of \cref{tab::posefree} predicts the poses and intrinsics of the input views, renders at the ground-truth target cameras and aligns the scale of every scene to the ground truth.
Its PSNR is below that of the protocol of \cref{tab::main} on both indices.
The effect of ground-truth intrinsics was tested in this setting only.
With the ground-truth intrinsics of the input views in place of the predicted ones, PSNR falls below that of this row, to $17.160$ and $16.536$\dB on the complete indices.
In training, a sample without the given cameras places the Gaussians with the predicted intrinsics.

\newcommand{\incellit}[1]{\includegraphics[width=0.096\textwidth]{cells/#1}}
\begin{figure*}[t]
  \centering
  \small
\setlength{\tabcolsep}{0.5pt}\renewcommand{\arraystretch}{0.8}
\begin{tabular}{w{c}{0.096\textwidth}w{c}{0.096\textwidth}w{c}{0.096\textwidth}w{c}{0.096\textwidth}w{c}{0.096\textwidth}w{c}{0.096\textwidth}w{c}{0.096\textwidth}w{c}{0.096\textwidth}w{c}{0.096\textwidth}w{c}{0.096\textwidth}}
No step & App.\ 1 & App.\ 2 & App.\ 3 & Shape 1 & Shape 2 & Shape 3 & Shape 4 & Shape 5 & GT \\
\multicolumn{10}{@{}c@{}}{\includegraphics{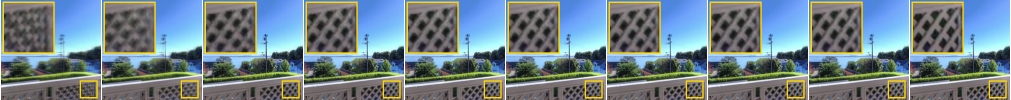}} \\
 22.97 & 28.06 & 32.09 & 33.99 & 35.04 & 35.62 & 36.09 & 36.32 & 36.45 & \\
\multicolumn{10}{@{}c@{}}{\includegraphics{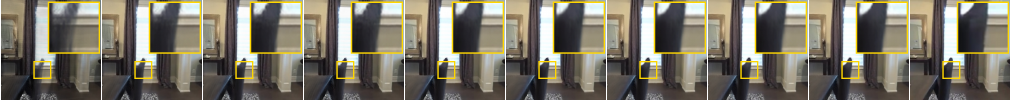}} \\
 17.88 & 27.39 & 30.71 & 29.03 & 32.65 & 32.91 & 33.96 & 34.41 & 34.70 & \\
\multicolumn{10}{@{}c@{}}{\includegraphics{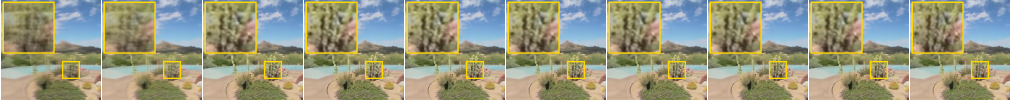}} \\
 25.95 & 30.99 & 34.56 & 35.04 & 36.60 & 37.19 & 37.47 & 37.65 & 37.73 & \\
\end{tabular}
  \caption{\textbf{Renderings after every step.} One target view of three scenes of the narrow protocol before any step, after each of the $K_a$ appearance steps and each of the $K_s$ shape steps of the final model, and in the ground truth.
  The PSNR of this view is beneath each rendering.}
  \label{fig::iters}
\end{figure*}

\begin{table}[!tb]\centering\footnotesize
\setlength{\tabcolsep}{5pt}
\caption{\textbf{Rotation noise on the input poses.} PSNR$\uparrow$ of the final model on the complete narrow protocol when every ground-truth input pose but the first is rotated by the given angle about a random axis.
The columns at $\poseYonoNarrow^{\circ}$ and $\poseOursNarrow^{\circ}$ are the mean rotation errors of the poses that YoNoSplat and \method predict, and the column at $0^{\circ}$ equals the entry of \cref{tab::main}.}
\label{tab::posesens}
\begin{tabular}{lcccccc}
\toprule
Noise & $0.00$ & $0.25$ & $0.50$ & $\poseYonoNarrow$ & $\poseOursNarrow$ & $1.00$\\
\midrule
PSNR$\uparrow$ & \mainNarrowOurs & 30.289 & 27.001 & 26.184 & 24.864 & 23.888\\
\bottomrule
\end{tabular}
\end{table}

\paragraph{Pose error.} On the complete narrow index the mean rotation error of the predicted input poses is $\poseOursNarrow^{\circ}$ for \method and $\poseYonoNarrow^{\circ}$ for YoNoSplat, with medians of $0.578^{\circ}$ and $0.436^{\circ}$.
On the complete wide index the means are $1.438^{\circ}$ and $0.929^{\circ}$ and the medians $1.045^{\circ}$ and $0.666^{\circ}$.
The normalized translation error is $0.1042$ for \method and $0.0720$ for YoNoSplat on the narrow protocol, and $0.0456$ and $0.0313$ on the wide protocol.
\Cref{tab::posesens} adds rotation noise to the ground-truth input poses of the final model and keeps everything else, including the ground-truth target cameras.
PSNR falls with every increase of the noise, from the smallest noise of $0.25^{\circ}$ on.
At the rotation error of \method it is lower than at that of YoNoSplat, and lower than the pose-free row of \method in \cref{tab::main}.
The noise test perturbs the input poses and keeps the ground-truth target cameras fixed.
The pose-free protocol estimates the target cameras together with the input cameras, so the two stay consistent.
\Cref{tab::posesens} thus measures the sensitivity of the rendering to an error of the input poses relative to the target cameras, and does not decompose the pose-free deficit.

\section{Analyses of the Refinement}
\label{sec::supp-design}

\begin{table}[tb]\centering\footnotesize
\setlength{\tabcolsep}{5pt}
\caption{\textbf{Refinement steps.} PSNR$\uparrow$ of the Gaussians of the final model after every refinement step at inference, on the target views and on the input views rendered back, over the complete narrow and wide indices.
The first row is the prediction before any step, and the last row is the output of \method.}
\label{tab::steps}
\begin{tabular}{lcccc}
\toprule
 & \multicolumn{2}{c}{Narrow protocol} & \multicolumn{2}{c}{Wide protocol}\\
\cmidrule(lr){2-3}\cmidrule(lr){4-5}
Gaussians after & Target & Input & Target & Input\\
\midrule
No step & 29.731 & 29.693 & 24.450 & 25.512\\
\midrule
Appearance step 1 & 31.501 & 31.667 & 25.213 & 26.668\\
Appearance step 2 & 31.969 & 32.237 & 25.379 & 26.931\\
Appearance step 3 & 31.813 & 32.082 & 25.260 & 26.738\\
\midrule
Shape step 1 & 32.807 & 33.392 & 26.081 & 28.346\\
Shape step 2 & 33.233 & 34.024 & 26.422 & 29.111\\
Shape step 3 & 33.460 & 34.376 & 26.592 & 29.492\\
Shape step 4 & 33.598 & 34.599 & 26.693 & 29.729\\
Shape step 5 & \cellcolor{best}\textbf{33.683} & \cellcolor{best}\textbf{34.744} & \cellcolor{best}\textbf{26.755} & \cellcolor{best}\textbf{29.888}\\
\bottomrule
\end{tabular}
\end{table}

\paragraph{Refinement steps.} \Cref{tab::steps} follows the Gaussians of the final model through the steps at inference, with fixed weights.
The steps read only the input views, and PSNR rises on the target views as well as on the input views on both protocols.
The appearance steps raise it from $29.731$ to $31.813$\dB on the narrow target views, and every column peaks at the second of them.
Every shape step raises PSNR in every column, and each gains less than the one before.

\paragraph{Per-step renderings.}\label{sec::supp-iters} \Cref{fig::iters} renders the Gaussians of the final model after every step.
Training supervises every iterate with \cref{eq::deepsup} and has no term on the order of the iterates.

\begin{table}[tb]\centering\footnotesize
\setlength{\tabcolsep}{6pt}
\caption{\textbf{Per-scene optimization by parameter group.} PSNR$\uparrow$ on the target views after $1500$ Adam~\cite{kingma2015adam} steps on the input views that update the ticked groups and keep the others as predicted.
Mean over $8$ scenes of the narrow protocol, starting from a base predictor of an earlier training run, our model without the shape head and the refinement, which scores $\groupsBase$.
The groups are centers $\bm{\mu}$, scale $\bm{s}$, rotation $\bm{R}$, opacity $\alpha$ and color $\bm{c}$, and DoF counts the optimized degrees of freedom per Gaussian.
The learning rate is $3\times10^{-3}$, and $10^{-1}$ for color alone.}
\label{tab::groups}
\begin{tabular}{ccccccc}
\toprule
$\bm{\mu}$ & $\bm{s}$ & $\bm{R}$ & $\alpha$ & $\bm{c}$ & DoF & PSNR$\uparrow$\\
\midrule
 & & & \cmark & & 1 & 31.401\\
 & & & & \cmark & 12 & \cellcolor{second}32.852\\
 & & & \cmark & \cmark & 13 & 32.767\\
 & & \cmark & & & 4 & 28.849\\
 & \cmark & & & & 3 & 32.650\\
 & \cmark & \cmark & & & 7 & \cellcolor{best}\textbf{\groupsShape}\\
\midrule
\cmark & \cmark & \cmark & \cmark & \cmark & 23 & \groupsAll\\
 & \cmark & \cmark & \cmark & \cmark & 20 & \cellcolor{best}\textbf{\groupsNoCentres}\\
\bottomrule
\end{tabular}
\end{table}

\paragraph{Parameter groups.} \Cref{tab::groups} optimizes one group of Gaussian parameters at a time on the input views and keeps the others as predicted.
It is measured on eight scenes of the narrow protocol at $224\times224$ pixels, from the base predictor, which comes from an earlier training run than the first stage of \cref{tab::stageeval}.
Scale and rotation reach a higher PSNR than opacity and color with fewer degrees of freedom.
They come close to the last two rows, which optimize all groups with and without the centers.
Optimizing the centers as well does not raise PSNR.
Within shape, scale alone gives most of the gain and rotation alone little.
On the same eight scenes, \method scores $34.379$\dB with its eight steps.
This lies between the last two rows, which take $1500$ steps from the earlier base predictor.
The base predictor is a model without the shape head and the refinement, which scores $27.130$\dB on $200$ scenes of the narrow protocol.
Across the four runs behind the table, scale and rotation were measured three times, at $34.207$, $34.210$ and $34.219$\dB, and all groups but the centers twice, at $34.505$ and $34.488$\dB.
The learning rate was swept over $3\times10^{-4}$, $3\times10^{-3}$, $3\times10^{-2}$ and $10^{-1}$ for color, for opacity, and for scale with rotation.
The best rate is $10^{-1}$ for color and $3\times10^{-3}$ for the other two, and the remaining rows use $3\times10^{-3}$ without a sweep.
The row of opacity with color is not tuned in this way.
At rates of $0.3$ and $1.0$, color alone reaches $32.765$ and $24.794$\dB and opacity with color $31.933$ and $17.417$\dB.
Over all these settings opacity and color reach at most $32.852$\dB, and scale with rotation reaches $\groupsShape$\dB.
With all groups optimized, PSNR is $28.101$, $29.379$, $30.511$, $32.407$, $33.657$, $34.138$ and $\groupsAll$\dB after $1$, $3$, $8$, $30$, $100$, $300$ and $1500$ Adam steps.
In \cref{tab::groups} the optimization with the centers frozen ends above that of all groups.

\begin{table}[!tb]\centering\footnotesize
\setlength{\tabcolsep}{4pt}
\caption{\textbf{Per-scene optimization by group on an intermediate model.} PSNR$\uparrow$, SSIM$\uparrow$, LPIPS$\downarrow$ and the high-frequency ratio HF on the target views after $1500$ Adam steps on the input views, as a mean over the eight scenes of \cref{tab::groups}.
The model already has the shape head and the shape steps, and every row uses a learning rate of $3\times10^{-3}$.}
\label{tab::lpipsgroups}
\begin{tabular}{lcccc}
\toprule
Optimized groups & PSNR$\uparrow$ & SSIM$\uparrow$ & LPIPS$\downarrow$ & HF\\
\midrule
None & 29.745 & 0.9357 & 0.0852 & 0.857\\
Opacity and color & 32.975 & 0.9590 & 0.0860 & 1.048\\
Scale and rotation & 34.275 & 0.9694 & 0.0668 & 0.902\\
All but the centers & 34.497 & 0.9710 & 0.0542 & 0.991\\
\bottomrule
\end{tabular}
\end{table}

\paragraph{Perceptual quality by group.} \Cref{tab::lpipsgroups} repeats the optimization by group on an intermediate model that already has the shape head and the shape steps.
This model is neither the base predictor of \cref{tab::groups} nor the final model.
The table uses the same eight scenes and a learning rate of $3\times10^{-3}$ in every row, so color is not at its tuned rate.
Optimizing opacity and color raises PSNR and leaves LPIPS slightly worse than before the optimization.
Optimizing scale and rotation lowers LPIPS, and optimizing all groups but the centers lowers it further.
The high-frequency ratio divides the high-frequency energy of a rendering by that of the ground truth.
With opacity and color it rises above $1$, the value of the ground truth, while LPIPS does not improve.
On the wide protocol AnySplat has a ratio of $1.028$ and the lowest PSNR of \cref{tab::main}, so the ratio is not a quality criterion on its own.

\paragraph{Shape statistics.} The statistics of the shape compare the base predictor with the result of $100$ Adam steps on scale and rotation at a learning rate of $3\times10^{-3}$, on the same eight scenes.
They count the Gaussians with a predicted opacity above $0.05$, which are $48.7\%$ of all Gaussians.
After $1500$ steps the scales collapse toward their lower bound and the median axis ratio is $1635$.
The ratio of the longest to the shortest axis rises from $4.2$ to $126.5$ over the $100$ steps.
The ratio is the per-scene median over the counted Gaussians, averaged over the scenes.
The per-scene medians after the steps range from $2.4$ to $528.0$.
In two scenes the median stays near the prediction and moves from $3.6$ to $2.4$ and from $3.9$ to $5.3$.
A ratio of $126.5$ is a log ratio of $4.8$, below the bound of $8$ that \cref{eq::aniso} puts on the change of a log axis ratio.
The median scale, averaged in the same way, changes from $0.02308$ to $0.01984$.
The angle between the shortest axis before and after the steps has quartiles of $1.3^{\circ}$, $6.9^{\circ}$ and $89.2^{\circ}$ over all counted Gaussians.
Three of the eight scenes have a median of $84^{\circ}$ to $87^{\circ}$.
The predicted direction of the shortest axis is kept at the median and not at the upper quartile.
The angle between the shortest axis after the steps and the normal of the predicted depth has quartiles of $36.7^{\circ}$, $58.6^{\circ}$ and $76.9^{\circ}$, and a random direction gives an expected $57.3^{\circ}$.
We therefore give the shape head no prior that aligns the shortest axis with this normal.

\paragraph{Predictability of the scale correction.} Three probes test whether the correction of the scale can be predicted from the inputs of the shape head.
They use an intermediate model on scenes of the narrow protocol.
It is the base predictor with a shape head trained alone, all other weights frozen.
The first probe is a ridge regression onto the log-scale correction found by $100$ Adam steps on the scale alone.
Its input is the $262$ input channels of the shape head, the $256$ of $\bm{F}$ and the six Sobel responses.
On $16$ scenes with $4{,}000$ samples each, the held-out $R^2$ is $0.0598$ at the best ridge weight.
A second run on twelve scenes with $42{,}000$ samples gives $0.0647$, and $0.0828$ and $0.0843$ once the backbone layers $4$ and $11$ are added to the input.
A multilayer perceptron on the same input gives $0.0544$.
The second probe trains the weights of the shape head on $2$ to $16$ scenes for $1500$ Adam steps with the loss on the input views.
It measures the change of PSNR on the target views of these scenes and of eight other scenes.
With $2$, $4$, $8$ and $16$ training scenes PSNR rises on the training scenes and falls on the other scenes.
The third probe fits the weights of the shape head to a single scene for $1500$ Adam steps, on five scenes that start at a mean of $24.509$\dB.
It reaches $31.448$\dB.
Optimizing the log-anisotropy $\bm{\delta}$ of every Gaussian directly, without rotation, reaches $30.693$\dB.
The two fits differ in their degrees of freedom, since the head also outputs a rotation increment, and neither is the refinement of \cref{eq::update}.
A linear map of the inputs of the head explains little of the variance of the scale correction, and a head fitted to $2$ to $16$ scenes lowers PSNR on other scenes.
A fit on the input views of one scene raises PSNR on its target views.

\paragraph{Global rescaling.} Multiplying all scales of the base predictor by one factor of $0.85$, $1.0$ and $1.3$ gives a PSNR of $26.178$, $\groupsBase$ and $27.908$\dB on the scenes of \cref{tab::groups}.
The factor of $1.3$ gives the highest of the three values, which stays far below the optimization of the scale of every Gaussian in \cref{tab::groups}.
The projector $\bm{P}$ of \cref{eq::aniso} removes a common factor from the correction, so the shape head and the shape steps do not apply this rescaling.

\paragraph{Gate.} The shape head has no learned gate on its output, since a gate initialized to zero in front of a randomly initialized projection read $0.019$ after $15{,}000$ training steps in an earlier module.

\paragraph{Adam steps and an earlier update.} Two earlier measurements relate gradient steps to PSNR, and neither uses the final model.
The first optimizes all Gaussian parameters, centers included, with Adam at a learning rate of $3\times10^{-3}$ on eight scenes of the wide protocol.
It starts from an early base predictor after $15{,}000$ training steps, which scores $21.50$\dB on these scenes.
PSNR is $22.142$, $22.559$, $23.285$, $24.744$, $25.926$, $26.258$ and $26.327$\dB after $1$, $3$, $8$, $30$, $100$, $300$ and $1500$ Adam steps.
On eight scenes of the narrow protocol the same predictor starts at $27.97$\dB and reaches $28.765$, $29.383$, $30.387$, $32.047$, $33.977$, $34.658$ and $34.623$\dB after the same numbers of steps.
The second trains two models for $15{,}000$ training steps that differ only in three steps of a learned update without hidden state.
This update acts on all Gaussian parameters and precedes the design of \cref{eq::update}.
On $32$ scenes of the wide protocol, the model with the update scores $22.453$, $23.402$ and $23.780$\dB after $5{,}000$, $10{,}000$ and $14{,}999$ training steps, and the other $20.670$, $21.862$ and $22.060$\dB.
On $200$ scenes after training it scores $23.392$\dB on the wide and $28.707$\dB on the narrow protocol, and the other model scores $21.756$ and $26.185$\dB.

\paragraph{Scene-level comparison.}\label{sec::supp-lpips} On the narrow protocol DepthSplat and VolSplat are the closest to \method on every metric of \cref{tab::main}.
Scene by scene, \method has a lower LPIPS than DepthSplat on $5{,}997$ of the $7{,}128$ scenes and than VolSplat on $6{,}056$.
It has the higher PSNR on $6{,}494$ and $6{,}583$ scenes and the higher SSIM on $6{,}445$ and $6{,}495$.

\section{Implementation Details}
\label{sec::supp-impl}

\paragraph{Backbone and dense head.} The backbone is VGGT-$\Omega$~\cite{wang2026vggtomega} with a patch size of $16$, so a view of $224\times224$ pixels gives $196$ patch tokens, one camera token and $16$ register tokens.
The camera head predicts a translation, a quaternion and two fields of view for every view.
The dense head fuses four layers of the backbone into a map of $256$ channels at a quarter of the image resolution.
A $1\times1$ convolution and a pixel shuffle by $4$ give the logarithm of the depth, which is clamped to $[-12,9]$ before the exponential.
The feature map $\bm{F}$ is computed from this fused map.
A zero-initialized convolutional embedding of the input image is added to it, a transposed convolution doubles the resolution, and a residual branch of six $3\times3$ convolutions with GELU and a zero-initialized $3\times3$ convolution is added.
A $1\times1$ convolution to $92$ channels and a pixel shuffle by $2$ then give $23$ values per pixel.
The backbone runs in bfloat16 precision.

\paragraph{Gaussian parameters.} The $23$ values are one for the opacity, three for the raw scale $\bm{\rho}$, four for the quaternion, twelve for the color and three for the offset of the center.
The opacity is the sigmoid of its value, the base scale follows \cref{eq::base}, and the quaternion is normalized.
The color uses spherical harmonics of degree $1$, and its constant term is the color of the input pixel plus a predicted residual.
The offset of the center is $0.05\tanh(\cdot)$ in world units and is added to the unprojected pixel.
The weights of the $1\times1$ convolution that outputs these values are initialized to zero, with a bias of $-2$ for the opacity and of $4$ for the raw scale.
The initial output is therefore a colored point cloud at the predicted depth, with an opacity of $0.12$ and a scale of about $0.004$.
Six views of $224\times224$ pixels give $301{,}056$ Gaussians.
The rasterizer is that of the gsplat library~\cite{ye2025gsplat} in its classic mode, with a black background and a 2D dilation of $0.3$.

\paragraph{Shape head.} The input image is resized bilinearly to the resolution of $\bm{F}$, and fixed Sobel filters in the two image directions give six response maps, two per color channel.
A $3\times3$ convolution of $\bm{F}$ and a $3\times3$ convolution of the Sobel responses, both to $128$ channels, are summed and followed by a GELU, a second $3\times3$ convolution and a GELU.
A zero-initialized $1\times1$ convolution to $24$ channels and a pixel shuffle by $2$ give the six raw values of every pixel, which form $\bm{r}_0$.

\paragraph{Shape update.} The update $\mathcal{U}_\psi$ of \cref{eq::update} reads $\bm{r}_k$, $\hat{\bm{g}}_k$, $\Delta\bm{r}_{k-1}$ and $\bm{v}_k$, with six channels each, and the $256$ channels of $\bm{F}$.
Two $3\times3$ convolutions with GELU encode these $280$ channels into $128$, and a $3\times3$ convolution of $\bm{F}$ gives a conditioning map of $96$ channels.
Four transformer blocks then refine the encoding.
A $1\times1$ convolution to $64$ channels, a GELU and an $8\times8$ convolution with stride $8$ turn each view into $14\times14$ tokens of $768$ dimensions, and a learned position embedding is added.
The blocks are pre-norm blocks with $12$ attention heads and an MLP of ratio $4$.
The first and third attend over the tokens of one view, and the second and fourth over the tokens of all views.
A layer norm, a linear layer and a pixel shuffle by $8$ return $64$ channels at the resolution of $\bm{F}$, and a GELU and a zero-initialized $1\times1$ convolution map them to $128$ channels that are added to the encoding.
A convolutional gated recurrent unit with $3\times3$ gates and a hidden state of $128$ channels reads the encoding and the conditioning map.
A zero-initialized $1\times1$ convolution maps the hidden state to $\Delta\bm{r}_k$.
The update has about $36.73$ million parameters, $34.82$ million of them in the transformer blocks.
The root mean square that normalizes $\bm{g}_k$ is bounded from below by $10^{-8}$, and $\bm{v}_k$ has no bias correction.

\paragraph{Appearance update.} The appearance update has the same layers, transformer blocks included, with a hidden state of $192$ channels and about $38.96$ million parameters.
The $23$ raw outputs of the dense head and their gradients are pixel-unshuffled by $2$ into $92$ channels each, so the input has $4\times92+256=624$ channels.
The output of $92$ channels is pixel-shuffled back, and its channels for scale, rotation and center offset are set to zero.

\begin{table}[!tb]\centering\footnotesize
\setlength{\tabcolsep}{5pt}
\caption{\textbf{Terms of the training loss.} Weight of every term and the quantity it is applied to.
The two distillation weights are the initial values, which \cref{eq::balance} then adjusts.}
\label{tab::losses}
\begin{tabular}{lll}
\toprule
Term & Weight & Applied to\\
\midrule
Mean squared error & $1$ & target views, every iterate\\
LPIPS & $0.05$ & target views, every iterate\\
D-SSIM & $0.8$ in stage 4 & target views, every iterate\\
Opacity regularizer & $0.01$ & output $\theta_M$\\
Depth smoothness & $0.05$ & predicted depth\\
Pose & $0.1$ & camera head\\
Intrinsics & $0.5$ & camera head\\
Depth distillation & $0.312$ & predicted depth\\
Camera distillation & $0.931$ & camera head\\
\bottomrule
\end{tabular}
\end{table}

\paragraph{Losses.} \Cref{tab::losses} lists the terms of the training loss.
Only $\mathcal{L}_{\mathrm{rgb}}$ is applied to every iterate, with the weights of \cref{eq::deepsup}.
With $M=8$ steps, the iterate before any step has the weight $\gamma^{8}$ and the output the weight $1$.
D-SSIM uses a Gaussian window of $11\times11$ pixels with a standard deviation of $1.5$.
The opacity regularizer is the mean opacity of the output.
The pose loss is the mean relative rotation angle over all pairs of views plus a Huber loss on their relative translation with a factor of $100$.
The intrinsics loss compares the normalized focal lengths.
The frozen copy of VGGT-$\Omega$ runs on the same input views at every step.
Its depth is distilled with a scale- and shift-invariant $L_1$ loss on the inverse depth.
Its cameras are distilled with a Huber loss on the translation, an $L_1$ loss on the quaternion up to its sign, and an $L_1$ loss on the fields of view with a factor of $0.5$.

\paragraph{Distillation weights.} A gradient-balancing rule sets the two distillation weights, separately on the depth branch of the dense head and on the camera head.
Every $500$ training steps it measures two $L_1$ norms on these parameters.
They are $q_{\mathrm{dist}}$ for the gradient of the weighted distillation term and $q_{\mathrm{comp}}$ for the gradient of the competing terms.
The weight $w$ then moves halfway toward
\begin{equation}
\label{eq::balance}
w^{\star}=\min\Bigl(\max\Bigl(\eta\,\frac{q_{\mathrm{comp}}}{q_{\mathrm{dist}}/w},\,0.05\Bigr),\,20\Bigr),
\end{equation}
with $\eta=1.43$, the target ratio of the distillation gradient to the competing gradient.
The competing terms are the mean squared error, LPIPS, the opacity regularizer, the pose and intrinsics losses and the depth smoothness.
They exclude D-SSIM and the terms of \cref{eq::deepsup} for the earlier iterates.
Every stage restarts from the initial weights of \cref{tab::losses}.

\paragraph{Views and cameras in training.} A training sample draws $V$ uniformly from $\{2,\dots,12\}$.
Its frame span comes from one of the bands $15$ to $90$, $90$ to $190$ and $190$ to $280$, chosen with probabilities $0.6$, $0.2$ and $0.2$.
The span is multiplied by $(V-1)/5$ and limited to the length of the scene.
The input views are equally spaced over the span, and four target views are drawn uniformly inside it from the frames that are not inputs.
A sample keeps the given cameras with probability $1$ during the first $60\%$ of every stage.
This probability falls linearly to $0.9$ at $80\%$ of the stage and stays there.
Otherwise the Gaussians are placed, and the input views rendered, with the predicted poses and intrinsics.
In this case the depth is still scaled by the ratio of the given to the predicted camera distance, and the target views are rendered with the given cameras.

\begin{table}[!tb]\centering\footnotesize
\setlength{\tabcolsep}{2.0pt}
\caption{\textbf{Training schedule.} Steps, number of GH200 GPUs, duration, learning rate of the heads and updates, and trainable parameters in millions.
Stage $2$ adds the shape head with three shape steps, stage $3$ runs $K_s$ shape steps and adds the $K_a$ appearance steps, and stage $4$ adds D-SSIM and the transformer blocks of the two updates.}
\label{tab::stages}
\begin{tabular}{clccccc}
\toprule
Stage & Added & Steps & GPUs & Hours & Rate & Params.\\
\midrule
1 & base predictor & $100{,}000$ & $16$ & $15.8$ & $2.8\times10^{-4}$ & $1148.6$\\
2 & shape head, steps & $30{,}000$ & $16$ & $6.2$ & $8\times10^{-5}$ & $1151.0$\\
3 & appearance steps & $24{,}000$ & $16$ & $7.1$ & $8\times10^{-5}$ & $1155.1$\\
4 & D-SSIM, attention & $24{,}000$ & $4$ & $21.4$ & $8\times10^{-5}$ & $1224.8$\\
\midrule
 & Total & $178{,}000$ & & $50.5$ & & \\
\bottomrule
\end{tabular}
\end{table}

\paragraph{Stages and optimizer.} \Cref{tab::stages} lists the four training stages.
All four use the full training split of RealEstate10K at $224\times224$ pixels.
Every stage starts from the weights of the previous one with a new optimizer and schedule.
The optimizer is AdamW~\cite{loshchilov2019adamw} with $\beta_1=0.9$, $\beta_2=0.95$ and a weight decay of $0.05$.
The heads and the two updates use the learning rate of \cref{tab::stages}.
The last layer of the backbone uses this rate with a multiplier of $0.3$, and the rate decays by a factor of $0.92$ per layer toward the input.
The rate warms up linearly over $2{,}000$ training steps and then follows a cosine to a tenth of its peak.
Every training step accumulates the gradients of $16$ scenes and clips the gradient norm at $1.0$.
A training step is skipped if its gradient norm exceeds the running median by a factor of $20$.

\section{Protocols, Indices and Baselines}
\label{sec::supp-protocols}

\paragraph{RealEstate10K indices.} Both published indices fix six input views and eight target views for every scene, so all methods render the same targets from the same inputs.
The narrow index, released with VolSplat, has $7{,}128$ scenes, and its inputs span a median of $49$ frames, between $13$ and $49$.
The wide index, released with YoNoSplat, has $1{,}600$ scenes, and its inputs span a median of $278$ frames, between $199$ and $278$.
From the narrow to the wide protocol, the PSNR of every method in \cref{tab::main} falls.
It falls least for AnySplat, YoNoSplat and ReSplat, then for \method, and most for pixelSplat, MVSplat, TranSplat, DepthSplat and VolSplat.

\paragraph{View-count index.} The view-count index keeps the eight targets of the narrow index and adds input views to its six, so a larger $V$ only adds inputs.
The frame window of a narrow scene holds at most $50$ frames, of which eight are targets, so $V=42$ is the largest count.
Scenes whose window cannot hold $42$ inputs are left out, which keeps $6{,}444$ of the $7{,}128$ scenes at every count.
Below six views the inputs are subsets of the published six, on the same scenes and targets.

\paragraph{Repeated frames.}\label{sec::supp-repeat} Some published indices name a target frame that is also an input frame of the same scene.
The narrow index has no such frame among its $57{,}024$ target frames, and the view-count index none among its $51{,}552$.
The wide index repeats $293$ of its $12{,}800$ target frames, in $273$ of its $1{,}600$ scenes.
The two-view index of RealEstate10K repeats $583$ of $19{,}422$ target frames and that of ACID $135$ of $4{,}785$.
On the wide protocol \method scores $30.023$\dB on the repeated target frames, $26.678$\dB on the others and $\mainWideOurs$\dB on all of them.
All tables use the published indices unchanged.

\paragraph{DL3DV index.} Transfer to DL3DV uses the index of the ReSplat repository, with $140$ scenes of six input and eight target views whose inputs span a median of $59$ frames.
The images of one scene have half the size of the others in the benchmark download and the conversion script drops it, so every method is evaluated on the same $139$ scenes.

\paragraph{Baseline weights.} All baselines are re-measured in our setting with released weights.
pixelSplat, MVSplat, TranSplat and DepthSplat use the weights of the VolSplat release, which its authors describe as trained with the same input views as VolSplat.
VolSplat trains with six input views that span $20$ to $50$ frames.
The pixelSplat and TranSplat files hold a view embedding with six entries.
The training steps stored in the files are $20{,}000$ for pixelSplat, $40{,}000$ for TranSplat and MVSplat, $150{,}000$ for DepthSplat and $145{,}000$ for VolSplat.
The TranSplat file has $575$ tensors with the names of the pixelSplat baseline, among them its epipolar transformer, and is run with the pixelSplat code.
This encoder adds its view embeddings in a random order at every forward pass, so the values of pixelSplat and TranSplat vary slightly between runs.
The released TranSplat code covers two, three and four input views.
ReSplat uses its two-view RealEstate10K weights with two update iterations.
YoNoSplat uses its RealEstate10K weights, trained with $2$ to $32$ input views, and its DL3DV weights in the lower block of \cref{tab::zeroshot}.
AnySplat uses its released weights, whose training mixture of nine datasets contains DL3DV and ScanNet++ and not RealEstate10K.

\paragraph{Near and far planes.} Every baseline renders with the near and far planes of its release.
They are $0.5$ and $100$ for DepthSplat, $1$ and $100$ for MVSplat, $0.1$ and $1000$ for pixelSplat and TranSplat, and $0.01$ and $100$ for ReSplat.
Outside \cref{tab::calib}, six-view rows use $224\times224$ pixels.

\paragraph{Self-reconstruction check.} In our setting the input views are rendered back, and their PSNR is reported as the self-reconstruction.
It requires $20$\dB of a baseline wherever this value is measured, and it is not measured for AnySplat.
On DL3DV, MVSplat and TranSplat are below this value, at $18.216$ and $19.822$\dB, and \cref{tab::zeroshot} reports them with the check relaxed.
NoPoSplat~\cite{ye2025noposplat} releases weights for two and three input views only and is not evaluated.

\begin{table}[!tb]\centering\footnotesize
\setlength{\tabcolsep}{3.6pt}
\caption{\textbf{Calibration of the setting.} PSNR$\uparrow$ printed in the six-view column of the VolSplat paper for the upper block and in the ReSplat paper for the lower block, and the value re-measured in our setting.
The upper block uses six input views on $200$ scenes of the narrow protocol at $256\times256$ pixels, and the lower block the two-view indices of \cref{tab::twoview}.}
\label{tab::calib}
\begin{tabular}{llcc}
\toprule
Method & Setting & Printed & Measured\\
\midrule
pixelSplat & RealEstate10K, $V=6$ & $28.95$ & $28.396$\\
MVSplat & RealEstate10K, $V=6$ & $29.13$ & $29.294$\\
TranSplat & RealEstate10K, $V=6$ & $29.62$ & $29.616$\\
DepthSplat & RealEstate10K, $V=6$ & $30.52$ & $30.784$\\
\midrule
ReSplat & RealEstate10K, $V=2$ & $29.75$ & $30.613$\\
ReSplat & ACID, $V=2$ & $29.87$ & $31.285$\\
\bottomrule
\end{tabular}
\end{table}

\paragraph{Calibration of the setting.}\label{sec::supp-calib} \Cref{tab::calib} compares the PSNR measured in our setting with a printed value.
In the upper block pixelSplat is measured below its printed value, TranSplat matches its own, and MVSplat and DepthSplat are above theirs.
In the lower block ReSplat is measured above its printed values.
With the image resizing of its repository, ReSplat scores $29.749$\dB in our setting on the complete RealEstate10K index, so the higher value comes from the resizing of our setting.
On ACID the adapter also aligns the poses to the middle input view, which the repository does on RealEstate10K only.
Our setting accepts a difference of $1.5$\dB.
On the complete narrow index at $224\times224$ pixels, the four baselines of the upper block are above their printed values in \cref{tab::main}.
VolSplat scores $31.243$\dB there, and its printed value is $31.30$\dB.

\paragraph{ScanNet++ index.} The index for ScanNet++~\cite{yeshwanth2023scannetpp} is built in this work, on its validation set nvs\_sem\_val.
The DSLR frames of a scene are ordered along a greedy nearest-neighbor path over the camera positions.
Six frames at regular intervals of a window of this path are the inputs, and three of the remaining positions are the targets.
The rule is deterministic, reads no score and puts no frame in both roles.
The largest angle between the optical axes of two input cameras has a median of $143.1^{\circ}$ over the scenes, and the angle between the optical axis of a target and that of its nearest input has a median of $4.8^{\circ}$.
The test frames of the dataset are not used as targets.
Their median distance to the nearest input is $0.44$ of the mean distance between the input cameras, and that of the targets of our index is $0.10$.
With the test frames as targets every evaluated method scores below $16$\dB.
The DSLR images are wide-angle.
After the center crop to a square, the normalized focal length is $0.397$ to $0.538$ over $30$ scenes, a horizontal field of view of $85.8^{\circ}$ to $103.2^{\circ}$.
It is $0.751$ to $1.276$ on RealEstate10K and $0.753$ to $1.160$ on DL3DV.
A further crop to the field of view of RealEstate10K raises the self-reconstruction of DepthSplat from $20.852$ to $23.641$\dB, and that of YoNoSplat in its pose-free mode from $10.195$ to $13.817$\dB.
The ScanNet++ columns of \cref{tab::zeroshot} use the images without this crop.

\end{document}